\documentclass{article}

\usepackage{arxiv}

\usepackage{amssymb}
\usepackage{amssymb} 
\usepackage{booktabs}
\usepackage{times}
\usepackage{lineno}
\usepackage{xcolor}
\usepackage{graphicx} 
\usepackage{float} 
\usepackage{tabularx}
\usepackage[inline]{enumitem}
\usepackage{amsmath}
\usepackage{amsfonts}
\usepackage{bm}
\usepackage{algorithm}
\usepackage[noend]{algpseudocode}
\usepackage{textcomp}
\usepackage{eso-pic}
\usepackage{tablefootnote}
\usepackage{threeparttable}
\usepackage{xpatch}
\usepackage[font=small,skip=0pt]{caption}
\usepackage{booktabs} 
\usepackage{tablefootnote} 
\usepackage{array,etoolbox}
\preto\tabular{\setcounter{magicrownumbers}{0}}
\newcounter{magicrownumbers}
\def\rownumber{}

\usepackage[utf8]{inputenc} 
\usepackage[T1]{fontenc}    
\usepackage{hyperref}       
\usepackage{url}            
\usepackage{nicefrac}       
\usepackage{microtype}      
\usepackage{lipsum}		
\usepackage{natbib}
\usepackage{doi}

\title{GNN-CNN: An Efficient Hybrid Model of Convolutional and Graph Neural Networks for Text Representation}

\author{ \href{https://orcid.org/0009-0000-2763-3906}{\includegraphics[scale=0.06]{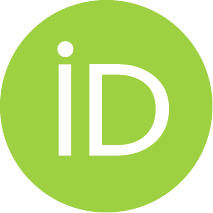}\hspace{1mm}Fardin Rastakhiz} \\
	\texttt{fardin.rastakhiz@gmail.com} 
}

\renewcommand{\shorttitle}{\textit{arXiv} Template}

\hypersetup{
pdftitle={GNN-CNN: An Efficient Hybrid Model of Convolutional and Graph Neural Networks for Text Representation},
pdfauthor={Fardin Rastakhiz},
pdfkeywords={First keyword, Second keyword, More},
}

\begin{document}
\maketitle

\begin{abstract}
	Time, cost, and energy efficiency are critical considerations in Deep-Learning (DL), particularly when processing long texts. Transformers, which represent the current state of the art, exhibit quadratic computational complexity relative to input length, making them inefficient for extended documents. This study introduces a novel model architecture that combines Graph Neural Networks (GNNs) and Convolutional Neural Networks (CNNs), integrated with a real-time, end-to-end graph generation mechanism. The model processes compact batches of character-level inputs without requiring padding or truncation. To enhance performance while maintaining high speed and efficiency, the model incorporates information from Large Language Models (LLMs), such as token embeddings and sentiment polarities, through efficient dictionary lookups. It captures local contextual patterns using CNNs, expands local receptive fields via lattice-based graph structures, and employs small-world graphs to aggregate document-level information. The generated graphs exhibit structural properties indicative of meaningful semantic organization, with an average clustering coefficient of approximately $0.45$ and an average shortest path length ranging between $4$ and $5$. The model is evaluated across multiple text classification tasks, including sentiment analysis and news categorization, and is compared against state-of-the-art models. Experimental results confirm the proposed model's efficiency and competitive performance. \href{https://doi.org/10.17632/d3cw4gyz85.3}{CNN-GNN Sources:} \citep{rastakhiz2025gnn}.
\end{abstract}

\keywords{ nlp \and text classification \and sparse transformer  \and convolutional neural network \and graph neural network}

\section{Introduction}
\label{intro}

\subsection{Importance}
\par Transitioning models from research to production introduces challenges in data collection, and high computational costs for scalability and resource efficiency \citep{paleyes2022challenges}. The escalating demands of time, cost, and energy in Deep Learning, especially with larger models, pose significant challenges to Artificial Intelligence (AI) businesses, both environmental and economic requiring researchers to seek efficient architectures and reduce raw floating-point operations (FLOPs) \citep{sharir2020cost, sevilla2022compute, strubell2020energy}. Deep learning models must be developed and deployed within realistic budgetary constraints to deliver timely business value \citep{benitez2018impact}. Factors such as dataset size, model complexity, and training duration significantly influence the required FLOPs \citep{amodei2018ai}. Consequently, the design and optimization of new deep learning models should explicitly address these factors. Various methods exist to achieve this, including model compression \citep{han2015deep}, knowledge distillation \citep{hinton2015distilling}, and hardware-aware training strategies \citep{hengyi2023hardware}, which aim to reduce computational complexity, accelerate training and inference, and minimize energy consumption. By prioritizing these considerations, researchers and practitioners can unlock the full potential of AI while ensuring its responsible and sustainable integration into real-world applications.

\par Natural Language Processing (NLP), at the intersection of AI and linguistics, empowers computers to understand and interact with human language, bridging the gap between human communication and machine understanding \citep{khurana2023natural}. The field encompasses a diverse range of text processing tasks, each with its own unique challenges and applications. These include fundamental tasks such as text classification \citep{reusens2024evaluating,fields2024survey} and named entity recognition \citep{hu2024deep}, as well as more complex tasks such as text embedding \citep{daneshfar2024elastic} and text generation \citep{li2024pre}. These capabilities underpin a wide array of real-world applications, from machine translation \citep{he2024exploring,li2024eliciting} and spam detection \citep{maurya2023deceptive,yadav2023machine} to more sophisticated uses in healthcare, finance, and education. 

\par To tackle these diverse NLP tasks, a plethora of model architectures has been developed, ranging from classical machine learning techniques like Term Frequency-Inverse Document Frequency (TF-IDF), Support Vector Machine (SVM), and Random Forests (RF) to deep learning models such as Recurrent Neural Networks (RNNs), CNNs, GNNs, Transformers, and numerous variants thereof.

\par Within the broader landscape of AI and NLP, deep learning has revolutionized NLP, enabling significant advancements in tasks such as machine translation, text summarization, sentiment analysis, and question answering \citep{young2018recent}. However, the success of deep learning in NLP has often come at the cost of increased model complexity and computational demands, exacerbating the challenges of time, cost, and energy efficiency discussed previously \citep{vaswani2017attention}. The trend toward larger language models, exemplified by models such as ChatGPT (Generative Pre-training Transformer), highlights this tension, offering impressive performance gains but requiring substantial resources for both training and deployment \citep{brown2020language}. This necessitates a focused investigation into methods for developing and deploying efficient deep learning models specifically tailored for NLP applications, balancing performance with practical constraints.


\par Deep learning's adaptability and parallel processing capabilities have driven its dominance in NLP, surpassing traditional methods by eliminating the need for complex feature engineering. While RNNs struggle with parallelism and CNNs are limited by local information processing, Transformers excel at capturing both local and global contexts; however, they face challenges with longer texts and input sizes. Studies emphasize evaluating model complexity against performance, noting that simpler models can rival more complex ones depending on the task and dataset. Models such as Fast-BERT \citep{liu2020fastbert}, a distilled version of BERT (Bidirectional Encoder Representations from Transformers), offer efficiency gains through distillation, although they introduce training complexities and retain quadratic complexity with respect to input size.

\par The quest for efficient NLP models has motivated significant research into architectures capable of effectively processing both local and global textual information while minimizing computational overhead. Processing local information, crucial for capturing immediate contextual cues, has been addressed through various architectural innovations. For instance, architectures such as QuickCharNet \citep{rastakhiz2024quickcharnet} and MobyDeep \citep{romero2022mobydeep} focus on computational constraints, leveraging techniques such as dilated and deformable convolutions \citep{chen2024review} and adapting local feature extraction strategies. Key components include convolutional layers, residual blocks, and global average pooling, all optimized for efficiency. Other approaches include standard Transformers and Heterogeneous Graph Attention Networks (HGAT) \citep{yang2021hgat}.

\par Capturing long-range dependencies and global context remains a central challenge. Near-linear complexity is a key consideration for deploying NLP models in real-world scenarios, especially when dealing with large volumes of text data. Transformer networks \citep{vaswani2017attention} have achieved state-of-the-art results on many NLP tasks, largely due to their attention mechanism's ability to model relationships between distant words in a sequence. However, the quadratic complexity of the standard attention mechanism with respect to sequence length has motivated the development of more efficient variants. 

\par Sparse Transformers, such as Exphormer \citep{shirzad2023exphormer}, and dynamic sparsity approaches \citep{jaszczur2021sparse} address this limitation by employing sparse attention patterns, reducing computational cost while retaining the ability to model long-range dependencies through graph-based structures. Dynamic sparsity provides a solution with reduced complexity of $O(n^{1.5})$ by selecting subsets of parameters per token. 

\par GNNs represent another approach to capturing global context, particularly in tasks where the input data can be naturally represented as a graph \citep{zhou2020graph}. By propagating information across graph nodes, GNNs can effectively model relationships between entities and capture global dependencies within the data.

\par Another method for optimizing large models is the use of Sparse Transformers, first introduced in \citep{child2019generating}. That paper proposed several enhancements to the Transformer architecture aimed at improving efficiency. These include reducing memory and time complexity relative to sequence length to $O(n \log n)$ through matrix factorization, and refining the model architecture to facilitate the training of deeper networks by restructuring residual blocks and optimizing weight initialization.

\par In addition to this work, many other papers have addressed the challenge of making Transformer models more efficient. The survey by \citep{tay2022efficient} provides a comprehensive overview of these approaches, collectively referred to as "X-formers", a term derived from model names such as Performer, Linformer, and Longformer.

\par Due to the sparsity of most graphs and the flexibility of graph structures, they can be effectively adapted to sparse Transformer architectures, opening up a research avenue for optimizing Transformer model design. The study by \citep{rampavsek2022recipe} proposed an architecture called General, Powerful, Scalable (GPS), which combines positional and structural encoding with local message passing and global attention mechanisms to enhance information aggregation across graphs. This model achieves linear complexity, enabling scalability to graphs with thousands of nodes. The GPS architecture demonstrates competitive performance compared to state-of-the-art models.

\par Another example of applying GNNs to improve Transformers is presented in \citep{shirzad2023exphormer}, which introduces Exphormer, a framework for building graph Transformers that allow nodes to attend to all other nodes. Exphormer utilizes virtual global nodes and expander graphs to enhance scalability and efficiency in graph learning tasks, enabling direct modeling of long-range interactions. By incorporating properties such as spectral expansion and pseudo-randomness, Exphormer reduces the number of training parameters, accelerates training speed, and maintains linear ($O(n)$) memory complexity proportional to the graph size. 

\par The paper by \citet{XIE2025112631} presents HCT-learn, a hybrid CNN-Transformer (HCT) framework that fuses spatial and temporal electroencephalogram (EEG) features to predict student learning outcomes. Initial convolutional layers extract local spatial patterns across multi-channel EEG, while a lightweight Transformer with sparse TopK self-attention captures long-range temporal dependencies at a reduced computational cost. By combining CNN-based spatial encoding with efficient global attention, the model achieves over 90\% accuracy.

\par Computing power is an important consideration in web development, where text content varies widely in length. Titles are short, comments are of medium length, and blog posts are long. Various tasks operate over web content, such as recommender systems, search engines, and sentiment analysis. The goal of this project is to develop a lightweight deep learning model capable of understanding both short and long texts, making it applicable to a range of tasks.
\par As the main contribution of this paper, we propose a new architecture that addresses these challenges by combining several solutions, including embedding and polarity injection, sparse transformers, efficient real-time graph generation, and a hybrid of convolution and attention mechanisms.

\begin{itemize}
    \item Developed a novel real-time graph generation layer to create more controlled and meaningful connections for sparse transformers.
    \item Removed the need for padding and truncation, enabling the model to handle texts of any length.
    \item Integrated additional architectural engineering techniques, such as embedding and polarity injection, to improve performance while maintaining high efficiency.
    \item Designed a lightweight preprocessing pipeline to maximize the practical benefits of the model.
\end{itemize}

Finally, the proposed model was compared with state-of-the-art models, and, using explainable AI methods, its features were visualized. The proposed model has memory and time complexity of $O(n)$ relative to text length.

\section{Methodology}
\label{method}
\par This section is structured to highlight the contributions of the proposed neural architecture by dividing it into three main subsections: data processing, real-time graph generation, and model design. First, we describe how raw inputs are curated and transformed to ensure both statistical integrity and reproducibility. Then, in the two subsequent subsections on graph generation and model architecture, we detail the architectural choices that distinguish our network from existing baselines, emphasizing both theoretical motivations and practical constraints.

\subsection{Data Processing}
\label{section:data_processing}
\par Data is a fundamental component of every deep learning model. Although one of the goals of deep learning is to minimize data engineering, there remains a need for some degree of data processing and, occasionally, feature extraction. However, the extracted features are generally not domain-specific and can be efficiently computed. In this section, we prepare three sets of token embeddings for future efficient integration into our model, outline the metadata extracted from raw texts for input, and analyze the time and space complexity of these practical processes.

\subsubsection{Prepare Metadata}
\textbf{Prepare token embeddings from LLMs:}
\par In our previous work \citet{rastakhiz2024quickcharnet}, we found that using the BERT tokenizer for URLs yields better results. However, this study explores a broader range of tasks involving natural language text, which follows more diverse linguistic structures. Therefore, three sets of embeddings were extracted from the following pre-trained models: DeBERTaV3-large with a vocabulary size of 128{,}100; OpenAI-GPT/tiktoken with 100{,}256 tokens; and SpaCy-large with 684{,}830 vocabulary items. The DeBERTa embeddings have 1{,}024 dimensions, GPT-3 embeddings have 3{,}072 dimensions, and SpaCy-large embeddings have 300 dimensions. 

\par According to benchmarks on Papers with Code across various datasets, DeBERTaV3 \citep{he2021debertav3} outperforms BERT \citep{devlin2019bert}, RoBERTa \citep{liu2019roberta}, and DeBERTa \citep{he2020deberta}. As a result, most of our experiments utilize DeBERTa embeddings. The DeBERTa tokenizer employs a subword tokenization strategy, breaking words into subword units to efficiently handle rare or out-of-vocabulary terms. Importantly, the tokenizer and the injected token embeddings must be compatible.

\par Using the extracted embeddings directly poses two major challenges: (1) the dimensionality varies across models, and (2) the embeddings are too large to be practical for lightweight models. To address these issues, Uniform Manifold Approximation and Projection (UMAP) was applied with 15 neighbors, reducing the dimensionality to either 64 or 128 components. The resulting reduced embeddings were then saved as compressed files for efficient storage and retrieval.

\par For downstream usage, the reduced embeddings were normalized using Min-Max normalization. These normalized embeddings were stored in a dictionary and integrated into the data loader, enabling dynamic construction of embedding tensors based on each document's tokens during training and inference.

\textbf{Prepare tokens' polarities:}
\par To enhance the model's ability to understand sentiment within texts, polarity and subjectivity scores were extracted for each token in the SpaCy vocabulary using the TextBlob package. Similar sentiment information was also obtained for ChatGPT and DeBERTaV3 tokens using the GPT-Chat API, and the results were saved in separate files. During preprocessing, these sentiment attributes were incorporated as token-level metadata and injected into the model to support downstream tasks such as sentiment classification.

\par A major challenge in this process lies in identifying an efficient method for integrating auxiliary sentiment information into the model architecture without significantly increasing computational complexity or compromising performance. This issue is further examined in the ablation study presented in Section \ref{section:ablation}.

\par \textbf{Character-token indices:}
\par An index tensor is constructed to aggregate the character embeddings corresponding to each token, utilizing a token-to-character index dictionary. The number of tokens in each document is also recorded and used for various purposes, including graph generation and global pooling across the document. To accurately define tokens, several tokenization methods—such as Tiktoken, SpaCy tokenizer, and BERT tokenizer—are evaluated for suitability.

\par \textbf{Input data structure:}

\begin{figure*}[!htbp]
  \centering
  \includegraphics[width=\textwidth]{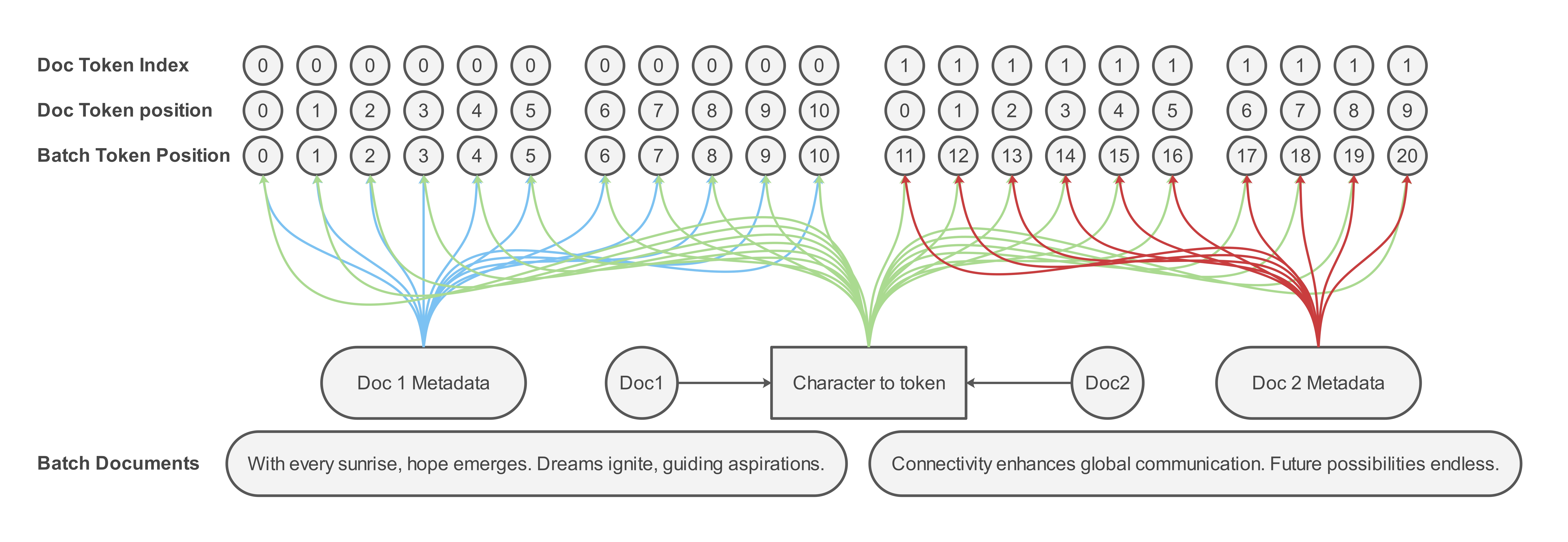}
  \caption{Token metadata required for graph construction} 
  \label{fig:token_graph_metadata} 
\end{figure*}

\par Some information is prepared within the data loaders. Figure \ref{fig:token_graph_metadata} illustrates several key components necessary for graph construction. In the batch data, no padding or truncation is applied. Instead, the processed data from documents are concatenated into a single, continuous sequence. In this structure, the character index of the first token in each document immediately follows the last character index of the preceding document, as illustrated in Figure \ref{fig:token_graph_metadata}. This batching strategy is referred to as the 'compact batch' approach in this paper. The remaining metadata components are described below.
\begin{itemize}
  \item \textbf{Character indices} are used to retrieve character-level embeddings within the model.
  \item \textbf{Character count per document} is required to construct compact batches by ensuring sequential alignment across documents.
  \item \textbf{Token positions} are utilized for positional encoding in compact batches.
  \item \textbf{Token count per document} serves multiple purposes, including graph construction and compact batch creation.
  \item \textbf{Token indices and token character counts} are used to aggregate character-level embeddings into token-level embeddings.
  \item \textbf{Cumulative token indices and compact batches} are generated by the data loader using each document's token count, token indices, and character counts. This enables the construction of a compact character sequence while preserving all relevant metadata.
  \item \textbf{Token polarity and subjectivity scores} are auxiliary sentiment features injected into the model.
  \item \textbf{Token embedding tensors} are constructed in the dataset using token indices and a precomputed token-to-embedding dictionary, and are passed to the model as input features.
  \item \textbf{Token sub-sampling probability tensors} are used to downweight frequently occurring tokens, such as stopwords or punctuation marks, during training.
\end{itemize}

\par
\begin{figure}[!h]
  \centering
  \includegraphics[width=\columnwidth]{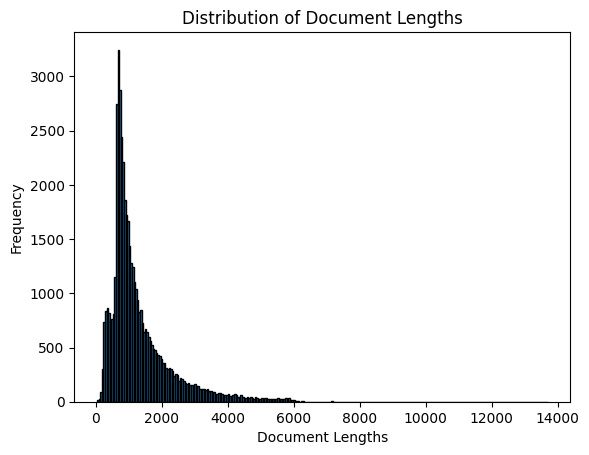}
  \caption{Document lengths in characters for the IMDB dataset}
  \label{fig:document_character_counts} 
\end{figure}

 One major challenge in using compact batches is the uncontrolled variation in document lengths. In some cases, a batch may consist primarily of large documents, while in others, it may include only short texts. Figure \ref{fig:document_character_counts} illustrates the distribution of document lengths in the Internet Movie Database (IMDB) dataset. As shown, a small number of very large documents are present, and their simultaneous inclusion in a batch may cause GPU memory overflow, becoming a critical bottleneck.

\par To address this issue, it is necessary to partition the dataset into batches with approximately equal total character counts while maintaining a consistent number of samples per batch. This requirement transforms the batching process into a $k$-way partitioning problem. To solve it, we adopt a modified version of the greedy number partitioning algorithm, as outlined in Algorithm \ref{alg:greed_k_way}. To ensure that the dataset size is divisible by the number of partitions $k$, additional dummy samples are introduced when necessary.

\begin{algorithm}[!h]
  \caption{Greedy \(k\)-way number partitioning with equal batch sizes}
  \label{alg:greed_k_way}
  \begin{algorithmic}[1]
    \Procedure{GreedyKPartition}{$L, k$}
    \State $n \gets \text{length of } L$
    \State $LA \gets \Call{DescendingSortArgs}{L}$
    \State $ranges \gets \Call{GenerateFixedSizeRanges}{0,n,k}$
    \If{$shuffle$ and $epoch > 0$}
    \State $LA \gets \Call{ShuffleInRanges}{LA, ranges}$
    \EndIf
    \State $L \gets L[LA]$
    \State $S[1 \dots k] \gets \text{array of } k \text{ empty lists}$
    \State $sum[1 \dots k] \gets \text{array of } k \text{ zeros}$
    \State $count[1 \dots k] \gets \text{array of } k \text{ zeros}$
    \For{$i = 1$ to $n$}
    \State $CI \gets \Call{Where}{count[j] < n/k}$
    \State $j \gets \text{partition index with smallest sum in CI}$
    \State $S[j] \gets S[j] \cup \{LA[i]\}$
    \State $sum[j] \gets sum[j] + L[i]$
    \State $count[j] \gets count[j] + 1$
    \EndFor
    \State Sort $S$ respect to the $sum$ in descending order
    \State \textbf{return} $S$
    \EndProcedure
  \end{algorithmic}
\end{algorithm}

\begin{algorithm}[!h]
   \caption{Token frequencies for sub-sampling}
  \label{alg:token_frequency}
    \begin{algorithmic}[1]
    \State \textbf{Input:} $documents$, $vocabularies$
    \State \textbf{Output:} $\mathbf{f}_{t}$
    \Statex
    
    \State $\mathbf{f}_{t} \gets$ mapping each token $t$ in $vocabularies$ to 1
    \State $temp \gets$ empty dictionary
    
    \For{each $doc$ in $documents$}
        \State $tokens\_list \gets tokenize(doc)$
        \State $new\_tokens \gets \{ \text{lower}(t) \mid t \in tokens\_list \}$
        \For{each token $t$ in $new\_tokens$}
            \If{$t \notin temp$}
                \State $temp[t] \gets 0$
            \EndIf
            \State $temp[t] \gets temp[t] + 1$
        \EndFor
    \EndFor
    
    \For{each $(k, v)$ in $\mathbf{f}_{t}$}
        \State $t_l \gets \text{lower}(k)$
        \If{$t_l \in temp$}
            \State $\mathbf{f}_{t}[k] \gets temp[t_l]$
        \Else
            \State $\mathbf{f}_{t}[k] \gets 1$
        \EndIf
    \EndFor
    \end{algorithmic}
    \footnotemark[1]{$\mathbf{f}_{t}$: All tokens frequencies.}
\end{algorithm}

\begin{algorithm}[!h]
   \caption{Linear token sub-sampling probabilities}
    \label{alg:linear_subsampling}
    \begin{algorithmic}[1]
    \State \textbf{Input:} $\mathbf{f}_{t,b}$
    \State \textbf{Output:} $\mathbf{p}_{ss,b}$
    \Statex
    \State $\mathbf{f}_{norm,t,b} = \frac{\mathbf{f}_{t,b}}{base\_corpus\_token\_count}$
    \Statex
    \State $\mathbf{p}_{ss,b} = \min\left(1, \frac{threshold}{\mathbf{f}_{norm,t,b}}\right)$ 
    \Statex
    \end{algorithmic}
    \footnotemark[1]{$\mathbf{f}_{t,b}$: Frequencies of batch tokens.} \newline
    \footnotemark[2]{$\mathbf{p}_{ss,b}$: sub-sampling probabilities of batch tokens.}
\end{algorithm}

\begin{algorithm}[!h]
    \caption{Experimental sigmoid sub-sampling probabilities}
    \label{alg:sigmoid_subsampling}
    \begin{algorithmic}[1]
    \State \textbf{Input:} $\mathbf{f}_{t,b}$
    \State \textbf{Output:} $\mathbf{p}_{ss,b}$
    \Statex
    \State $\mathbf{f}_{norm,t,b} = \frac{\mathbf{f}_{t,b}}{base\_corpus\_token\_count}$
    \Statex
    \State $\mathbf{p}_{ss,b} = 1-0.95.\sigma(0.05.((\frac{\mathbf{f}_{t,b}}{threshold})-90))$
    \Statex
    \end{algorithmic}
    \footnotemark[1]{$\mathbf{f}_{t,b}$: Frequencies of batch tokens.} \newline
    \footnotemark[2]{$\mathbf{p}_{ss,b}$: sub-sampling probabilities of batch tokens.}
\end{algorithm}

\par To introduce randomness into the greedy partitioning process described in Algorithm \ref{alg:greed_k_way}, two additional steps were incorporated. Specifically, at the beginning of each epoch, the data are shuffled to generate a new random order. This ensures that while the batch sizes remain controlled and balanced, the composition of each batch varies across epochs, promoting better generalization during training.

\par A well-known challenge in text processing is managing term frequency and document-term frequency, which has been addressed in various works through techniques such as TF-IDF. In this work, for the purpose of real-time processing, a term-frequency dictionary was constructed using a large corpus of text and saved for subsequent use. In the data loader, this dictionary is employed to compute sub-sampling probabilities for each term. These probabilities are then incorporated into the model's input data to reduce the influence of high-frequency terms and improve learning efficiency.

\subsubsection{Preprocessing Time and Space Complexity}

\begin{itemize}
    \item \textbf{One-hot encoding for labels:} Time complexity is $O(n_d \cdot n_c)$, where $n_d$ is the number of documents and $n_c$ is the number of classes.
    
    \item \textbf{WordPiece tokenization:} Time complexity is $O(n)$ and space complexity is $O(n + V)$, where $n$ is the length of a document and $V$ is the vocabulary size.
    
    \item \textbf{Repeat-interleave functions:} All have $O(n)$ time complexity, where $n$ is the length of the document.
    
    \item \textbf{Sub-sampling equation:} Dictionary generation is a one-time operation with time and space complexity of $O(n_d \cdot n + V)$. Subsequent token lookups in each document are performed in $O(n)$ time.
    
    \item \textbf{Greedy K-Way partitioning algorithm:} \\
    \textbf{Time complexity:} $O(n_d \log n_d)$ \\
    \textbf{Space complexity:} $O(n_d)$
\end{itemize}
Therefore, with considering one-time sub-sampling dictionary setup the preprocessing has $O(n_d.(n+n_c)+V+n_d.log(n_d))$ for time complexity and $O(n_d.(n+n_c)+V)$ for space complexity. without considering one-time sub-sampling dictionary setup and assuming that $n>>n_c$ and $n>>log(n_d)$, it has $O(n_d.n)$ for time complexity and $O(n_d.n+V)$ for space complexity.

\subsection{Real-time Graph Construction}
\label{section:graph_data_generation}
The real-time graph generator is integrated into the model; however, its primary role is to transform text-derived metadata into graph data. Accordingly, we describe it in this section.

\subsubsection{Important Graph Topological Features}
\par Graph topological features are critical for Graph Neural Networks (GNNs), as they capture the structural properties of the input data and inform the model about relationships and dependencies among nodes. The following topological features were considered in constructing the graphs used in our model:

\par \textit{Degree:} The degree of a node represents the number of edges connected to it. Nodes with high degrees often play a central role in the graph and may exhibit distinct characteristics compared to low-degree nodes. With the aid of sub-sampling during graph generation, high-frequency tokens that appear less frequently across the entire corpus tend to gain higher degrees in the constructed graph.

\par \textit{Centrality Measures:} Centrality metrics such as betweenness centrality, closeness centrality, and eigenvector centrality quantify the importance of nodes based on their position within the graph structure.

\par \textit{Clustering Coefficient:} The clustering coefficient measures the extent to which nodes in a graph tend to form tightly knit groups, indicating the presence of local structures such as communities or cliques. As shown by \citet{kaiser2008mean}, this metric is particularly informative in small-world networks and reveals the organizational patterns of the graph, including the influence of isolated nodes and leaves on classification and clustering. \citet{saramaki2007generalizations} further extend this concept by providing a generalized method to compute clustering coefficients for graphs with weighted edges.

\par \textit{Graph Diameter:} The graph diameter is defined as the longest shortest path between any pair of nodes. It offers insight into the overall connectivity and the efficiency of information propagation within the graph. As discussed by \citet{loukas2019graph}, the ability of GNNs to capture global information is influenced by the graph's diameter, as message passing in GNNs propagates information one hop per layer. Although deeper GNNs can theoretically capture long-range dependencies, in our proposed model, additional components such as convolutional layers also play a significant role in facilitating global information flow.

\par \textit{Graph Density:} Graph density is the ratio of the number of existing edges to the total number of possible edges. It reflects the level of connectivity and highlights whether the graph is sparse or dense. In contrast to fully connected architectures such as dense transformers, low-density graphs are computationally more efficient, often requiring significantly fewer FLOPs.

\par By incorporating these key graph topological features, GNNs can effectively capture the structural properties of the graph and learn meaningful representations for various graph-based tasks, including node classification, link prediction, and graph classification.


\subsubsection{Construction of the graph:}
\par Real-world networks often exhibit high average clustering and low average distances between node pairs. Several studies have addressed these properties and proposed solutions for various tasks, such as Exphormer \citep{shirzad2023exphormer}. In this study, both speed and performance are critical considerations. Therefore, a fast graph construction method that preserves these graph characteristics was integrated into the model to enable real-time execution. An efficient approach was also proposed for constructing both lattice and random graphs over the text data.

\begin{algorithm}[!h]
    \caption{Graph generation process}
    \label{alg:gen_graph}
    \begin{algorithmic}[1]
        \State \textbf{Input:} $\mathcal{X}_t, metadata$
        \State \textbf{Output:} $graph\_data$
        \State $R_b, L_b, \textbf{p}_b \gets calculate\_graph(\mathcal{X}_t, metadata)$    
        \State $E \gets create\_edges(R_b, L_b, \textbf{p}_b)$
        \State $E \gets  \text{subsample\_edges}(E, \mathbf{p}_{ss,b})$
        \State $data \gets \text{Data}(x=\mathcal{X}_t, edge\_index=E)$
        \State \textbf{Return:} $data$
    \end{algorithmic}
    \footnotemark[1]{$\textbf{p}_b$: token position over the batch} \newline
    \footnotemark[2]{$L_b$: lattice links over the batch} \newline
    \footnotemark[3]{$R_b$: random links over the batch}
    \footnotemark[3]{$E$: Edge indices}
\end{algorithm}

\begin{algorithm}[!h]
    \caption{Graph metadata calculation}
    \label{alg:calc_graph_metadata}
    \begin{algorithmic}[1]
        \State \textbf{Input:} $\mathcal{X}_t, metadata$
        \State \textbf{Output:} $R_b, L_b, \textbf{p}_b$
        \State Calculate the number of tokens in the batch $|\mathcal{B}_t|$, in each document of the batch $|\mathcal{D}_n|$, the lower bound $l_n$, and upper bound $u_n$ of each document set $\mathcal{D}_n$.
        \State The tokens' positions in the batch $\textbf{p}_b$ and their positions in their documents $\textbf{p}_n$, all starting from $0$:
        \begin{equation} \label{eq:token_batch_pos}
          \textbf{p}_b = \{0, 1, 2, \dots, |batch tokens|-1\}
        \end{equation}
        \begin{equation} \label{eq:token_doc_pos}
          \textbf{p}_n = \left\{i - l_n \mid i \in \textbf{p}_b \wedge t_i \in \mathcal{D}_n\right\}
        \end{equation}
        \State Identify each token's document, its number in the document, and its number in the batch.
        \State Generate a random number for each token concerning the corresponding token's document length $|\mathcal{D}_n|$. Then, select $K_1$ random nodes to create links like Figure \ref{fig:random_graph_nodes} using the following formula:
        \begin{equation} \label{eq:gen_random_number}
          \mathbf{r}_{v_i} = \text{int}((2*|\mathcal{D}_n| + 1) \cdot \text{rand}())
        \end{equation}
        \begin{equation} \label{eq:gen_random_nodes}
          R_b = (((\mathbf{r}_{v_i} \mod |\mathcal{D}_n|) + \textbf{p}_n + 1) \mod |\mathcal{D}_n|) + l_n
        \end{equation}
        \State Generate an array of $K_2$ regular nodes to create lattice links like Figure \ref{fig:lattice_graph_nodes} using the following formula:
        \begin{equation} \label{eq:gen_lattice_nodes}
          \textbf{l}_{v_i} = \{m, m + \text{{s}}, m + 2\cdot\text{{s}}, \ldots, m + k\cdot\text{{s}}\}
        \end{equation}
        \begin{equation} \label{eq:gen_lattice_nodes2}
          L_b = ((\textbf{l}_{v_i} + \textbf{p}_n) \mod |\mathcal{D}_n|) + l_n
        \end{equation}
    \end{algorithmic}
\end{algorithm}
    
\par
\begin{figure*}[!htb]
  \centering
  \includegraphics[width=\textwidth]{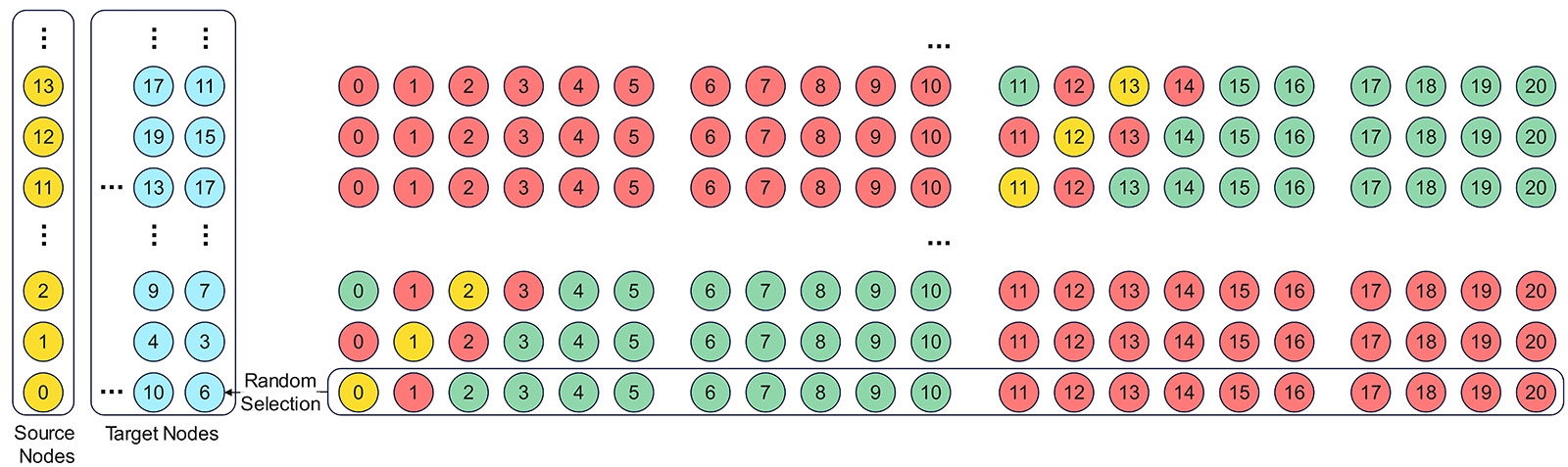}
  \caption{Random node selection for graph creation}
  \label{fig:random_graph_nodes} 
\end{figure*}

\begin{figure*}[!htb]
  \centering
  \includegraphics[width=\textwidth]{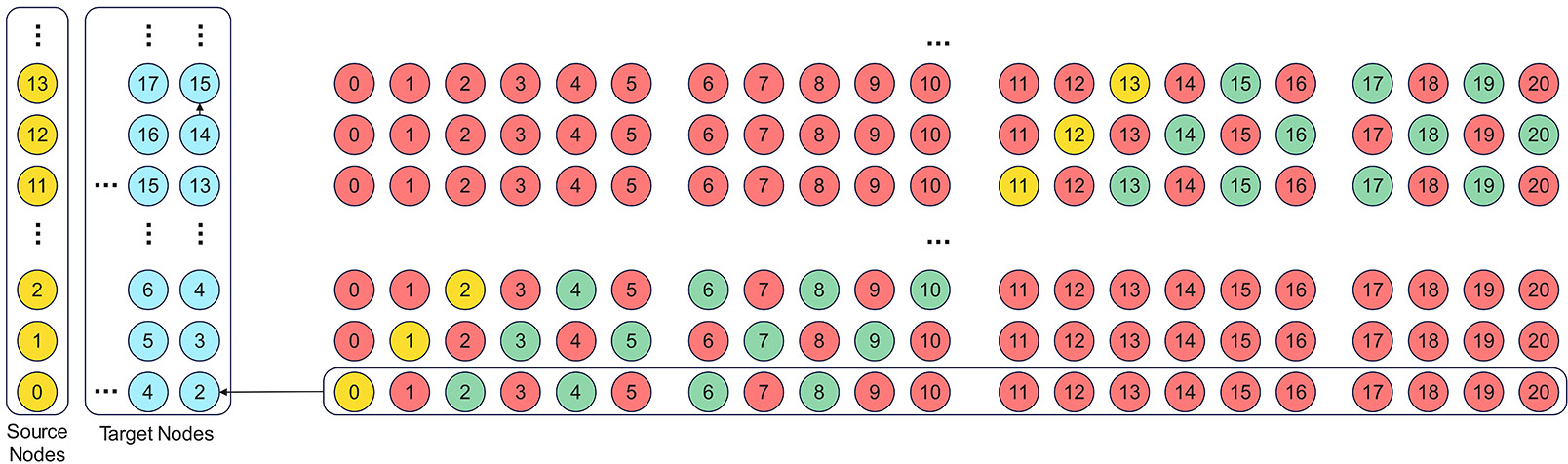}
    \caption{Regular node selection for lattice graph creation}
  \label{fig:lattice_graph_nodes} 
\end{figure*}

\begin{figure*}[!htb]
  \centering
  \includegraphics[width=\textwidth]{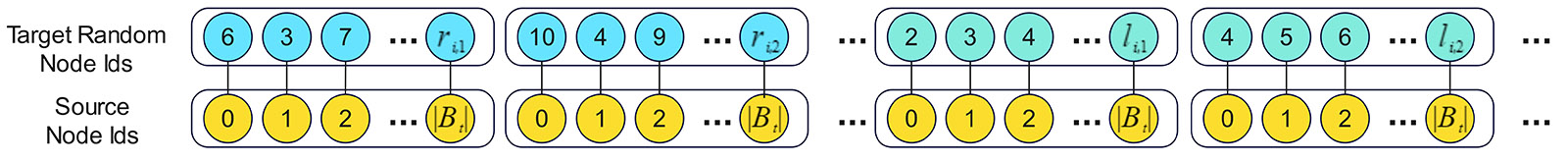}
  \caption{Edge creation by pairing source and target nodes}
  \label{fig:node_ids_arrangement} 
\end{figure*}

 To construct the graph, each token is treated as a node. Figures \ref{fig:random_graph_nodes}, \ref{fig:lattice_graph_nodes}, and \ref{fig:node_ids_arrangement} illustrate the final arrangement of random and lattice edges used to establish the graph structure. The random edges are regenerated in each iteration to prevent rapid overfitting and enhance the model's generalizability.

\begin{algorithm}[!h]
    \caption{sub-sampling and clustering edges}
    \label{alg:clustering_edge}
    \begin{algorithmic}[1]
        \State \textbf{Input:} $E$, $\mathbf{p}_{ss,b}$, $p_{keep}$
        \State \textbf{Output:} $E$
        \State $p_E=\text{sum}(\mathbf{p}_{ss,b}[E], axis=0)$
        \State $keep\_count=int(p_{keep} * \frac{edge\_count}{head})$
        \State $top\_k\_indices \gets \text{topk}(p_E, keep\_count, axis=0)$
        \State $E \gets E[top\_k\_indices]$
        \State \textbf{return} $E$
    \end{algorithmic}
    \footnotemark[1]{$P_{keep}$ is the ratio of the edge counts to keep.}
\end{algorithm}

\par The model's first Graph Attention Network (GAT) or attention layer operates on the constructed graph, producing both the output embeddings and the corresponding attention weights. These attention weights are then used to identify and replace the least important edges in the graph, followed by repeating the graph construction process. To evaluate the quality and effectiveness of the generated graphs, several sample features are presented below.

\begin{itemize}
\item For a document with 360 tokens: Density = 0.0551, Diameter = 4, Average Clustering = 0.463, Average Shortest Path = 2.55
\item For a document with 1{,}018 tokens: Density = 0.0196, Diameter = 4, Average Clustering = 0.450, Average Shortest Path = 2.94
\item For a document with 1{,}709 tokens: Density = 0.0117, Diameter = 5, Average Clustering = 0.447, Average Shortest Path = 3.17
\end{itemize}
The following graph construction settings were used to obtain the above results:
\begin{itemize}
\item Lattice begin distance: 2
\item Lattice step: 2
\item Number of lattice edges per node: 8
\item Number of random edges per node: 4
\end{itemize}

\par For longer texts with around 1{,}709 tokens (e.g., from the RT-2K movie reviews dataset), the resulting graph has a density of approximately 0.012. This observation, combined with other topological features mentioned earlier, demonstrates how the graph remains sparse even as text size increases, while still maintaining strong token connectivity and desirable structural properties for efficient learning.

\begin{figure*}[htb]
  \centering
  \includegraphics[width=1\textwidth]{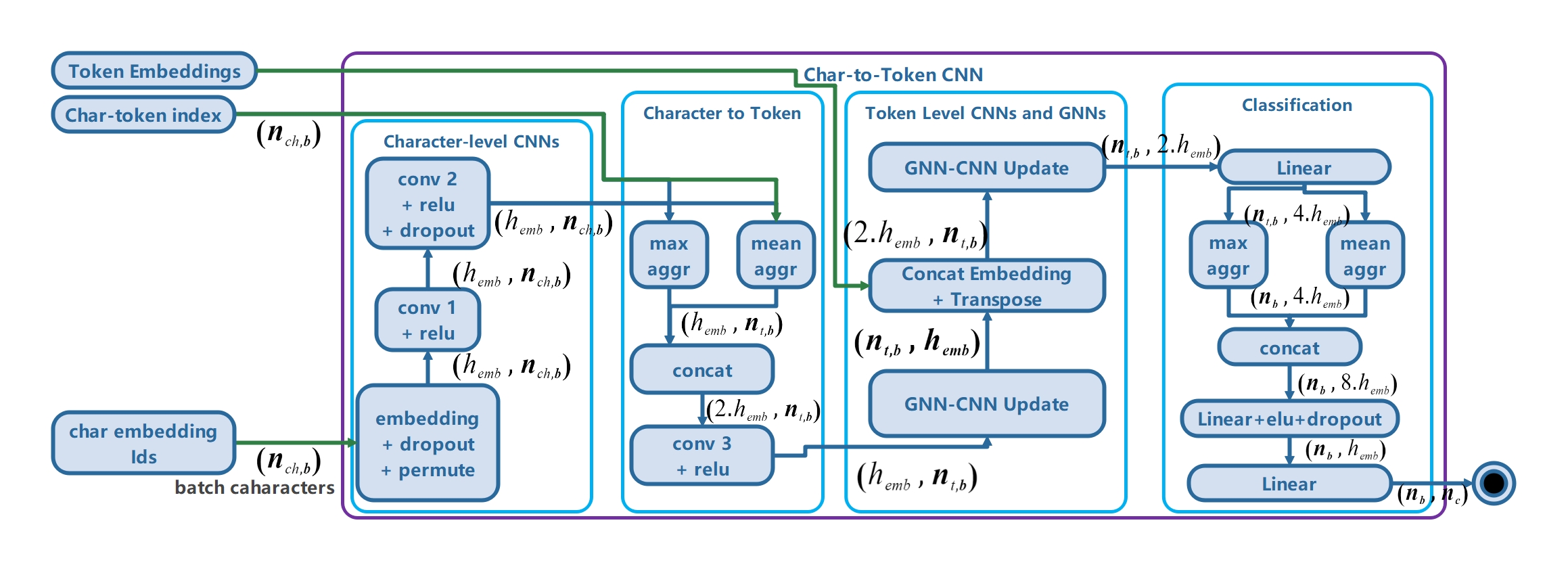}
  \caption{Diagram of the proposed model architecture. The model consists of several modules, some of which are illustrated in Figure \ref{fig:gnn_cnn_update} for the GNN-CNN update.}
  \label{fig:model_architecture} 
\end{figure*}

\begin{figure*}[htb]
  \centering
  \includegraphics[width=1\textwidth]{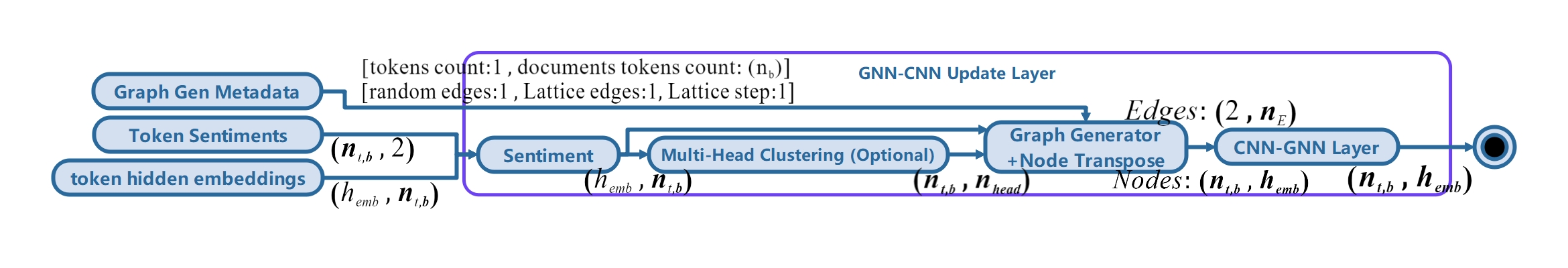}
  \caption{This module includes graph generation, sentiment incorporation (Figures \ref{fig:sentiment_type_1}, \ref{fig:sentiment_type_2}, \ref{fig:sentiment_type_3}), multi-head clustering (Figure \ref{fig:multi_head_clustering}), and graph processing using the CNN-GNN layer (Figure \ref{fig:cnn_gnn_layer}).}
  \label{fig:gnn_cnn_update} 
\end{figure*}

\begin{figure}[!htb]
  \centering
  \includegraphics[width=\columnwidth]{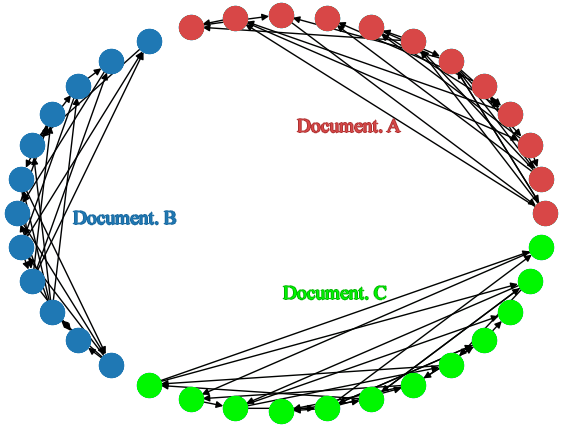}
\caption{Sample graph generated for a batch containing three texts, each with approximately 50 characters and about 12 tokens.}
  \label{fig:graph_generated} 
\end{figure}

\begin{figure}[!htb]
  \centering
  \includegraphics[width=\columnwidth]{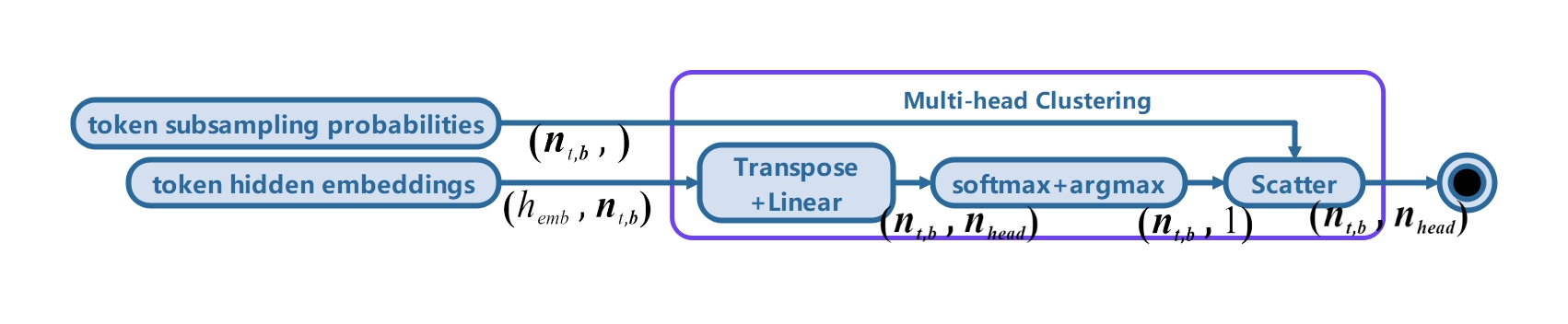}
  \caption{A clustering layer for multi-layer graph generation. This layer learns related concepts from previous layers and clusters nodes along with their metadata accordingly.}
  \label{fig:multi_head_clustering} 
\end{figure}

\subsubsection{Memory and Computational Complexity of the Real-Time Graph Generator}
\label{sec:preprocessing_complexity}

Since the number of edges per node is fixed, the total number of edges grows linearly with the number of tokens in the batch, denoted by $n$. Therefore, both the memory usage and computational complexity of the graph generation process are $\mathcal{O}(n)$.

\par \textit{Top-$k$ over edges using quicksort:}  The time complexity is $\mathcal{O}(n + k)$. Since $k$ is fixed and $k \ll n$, this simplifies to $\mathcal{O}(n)$.
\par \textit{Repeat interleave:} This operation has a time complexity of $\mathcal{O}(n)$.
\par \textit{'calculate\_graph' function:} This function includes generating a matrix of size $n$, performing repeat interleave, computing cumulative sums, and applying element-wise arithmetic operations. Each of these operations has at most $\mathcal{O}(n)$ complexity.
\par \textit{'fill\_lattice\_and\_random\_edges' function:} This function involves concatenating edge head indices with complexity $\mathcal{O}(n \cdot (k_1 + k_2))$, where $k_1$ is the number of lattice edges and $k_2$ is the number of random edges per node. Given that $k_1$ and $k_2$ are both fixed and satisfy $k_1 \ll n$, $k_2 \ll n$, the overall complexity simplifies to $\mathcal{O}(n)$.
\par Therefore, the overall time and space complexity of the graph generation process, relative to the input size $n$, is $\mathcal{O}(n)$. However, when accounting for all contributing variables, it can be more precisely described as $\mathcal{O}(n \cdot (k_1 + k_2))$.

\subsection{Model}
\label{section:model}

\par In this work, we propose an efficient model that combines key features from CNNs, GNNs, and Transformers to effectively handle long textual inputs. The architecture is designed to balance performance and efficiency, incorporating multiple components such as character-level input processing, token-level representation, semantic and sentiment injection, real-time graph construction, and a hybrid of GNNs, or modified attention mechanisms, and CNNs. Figure \ref{fig:model_architecture} illustrates the overall architecture of the proposed model, with detailed descriptions of each component presented in the following sections.

\par \textbf{Inputs:} As described in Section \ref{section:data_processing}, the model input consists of more than just character indices. It takes token information, their associated embeddings, polarity and subjectivity scores to enrich the token representations. Additionally, the model requires graph metadata to support the real-time graph construction process. In summary, all input components necessary for preparing the compact batch and related computations are detailed in Section \ref{section:data_processing} and Equation \ref{eq_chars}.

\begin{itemize}
  \item For document $i$ with $n_i$ character counts, the list of document characters is:
        \begin{equation} \label{eq_chars}
          \mathbf{x}_i=\left\langle x_i^1, x_i^2, \ldots, x_i^{n_i}\right\rangle
        \end{equation}
  \item In the following formulas for document $i$ with $m_i$ tokens, token $t_i^j$ is the $j$th token in $i$th document and $L$ is the number of characters in each tokens and $f$ is the character-token index array:
        \begin{gather*}
          l_i^j = len(t_i^j)                                                      \\
          L = \left\langle l_i^{t,1}, l_i^{t,2}, \ldots, l_i^{t,m_i}\right\rangle \\
          f((0, 1, \ldots, m_i), L)=\left\langle 0, \ldots, 0, 1, \ldots, 1, \ldots, m_i, \ldots, m_i \right\rangle
        \end{gather*}
\end{itemize}

\par \textbf{Embeddings:} The first layer of the model generates character-level embeddings using the provided character indices.

\begin{equation} \label{eq_embeddings}
  \mathbf{h}_{i} =\mbox{$Z\mathbf{x}_i$} 
\end{equation}
where $\mathbf{h}_{i}$ is the embedding vector of sample $\mathbf{x}_i$, and $Z$ is the embedding weight matrix.

\par \textbf{Character-Level Processing:} Character-level input offers several advantages over token-level input, such as a smaller vocabulary size—which leads to a smaller embedding layer \citep{kim2016character}, improved domain adaptation across various languages and specialized contexts, and reduced preprocessing requirements \citep{xuebyt52022, clark2022Canine}. However, character-level models also present challenges, including longer input sequences and a tendency to require deeper architectures \citep{clark2022Canine}. These issues are addressed in this paper as well as in \citep{rastakhiz2024quickcharnet, clark2022Canine}.

\par \textbf{Character-Level Processing:} Character-level input offers several advantages over token-level input, such as a smaller vocabulary size, resulting in a more compact embedding layer \citep{kim2016character}, better domain adaptation across various languages or specialized use cases, and reduced preprocessing requirements \citep{xuebyt52022, clark2022Canine}. However, it also presents challenges, including longer sequences and a tendency to require deeper models \citep{clark2022Canine}. These challenges are addressed in this paper, as well as in \citep{rastakhiz2024quickcharnet, clark2022Canine}.

\par As illustrated in Figure \ref{fig:model_architecture}, the model processes character-level input by applying character embeddings to two convolutional layers with Rectified Linear Unit (ReLU) activation functions, following the approach of \citep{zhang2015character} and \citep{rastakhiz2024quickcharnet}. A kernel size of 5 is used in the convolutional layers to capture broader contextual information. Unlike traditional approaches, no padding or truncation is required for input documents; instead, the input is treated as a compact sequence of all documents in the batch. Following this, aggregation functions, are applied over the character embeddings of each token. These functions are slightly modified to improve efficiency and compatibility with the new architecture, enabling the model to better capture the morphological structure of each token.

\par
\begin{figure}[!htbp]
  \centering
  \includegraphics[width=\columnwidth]{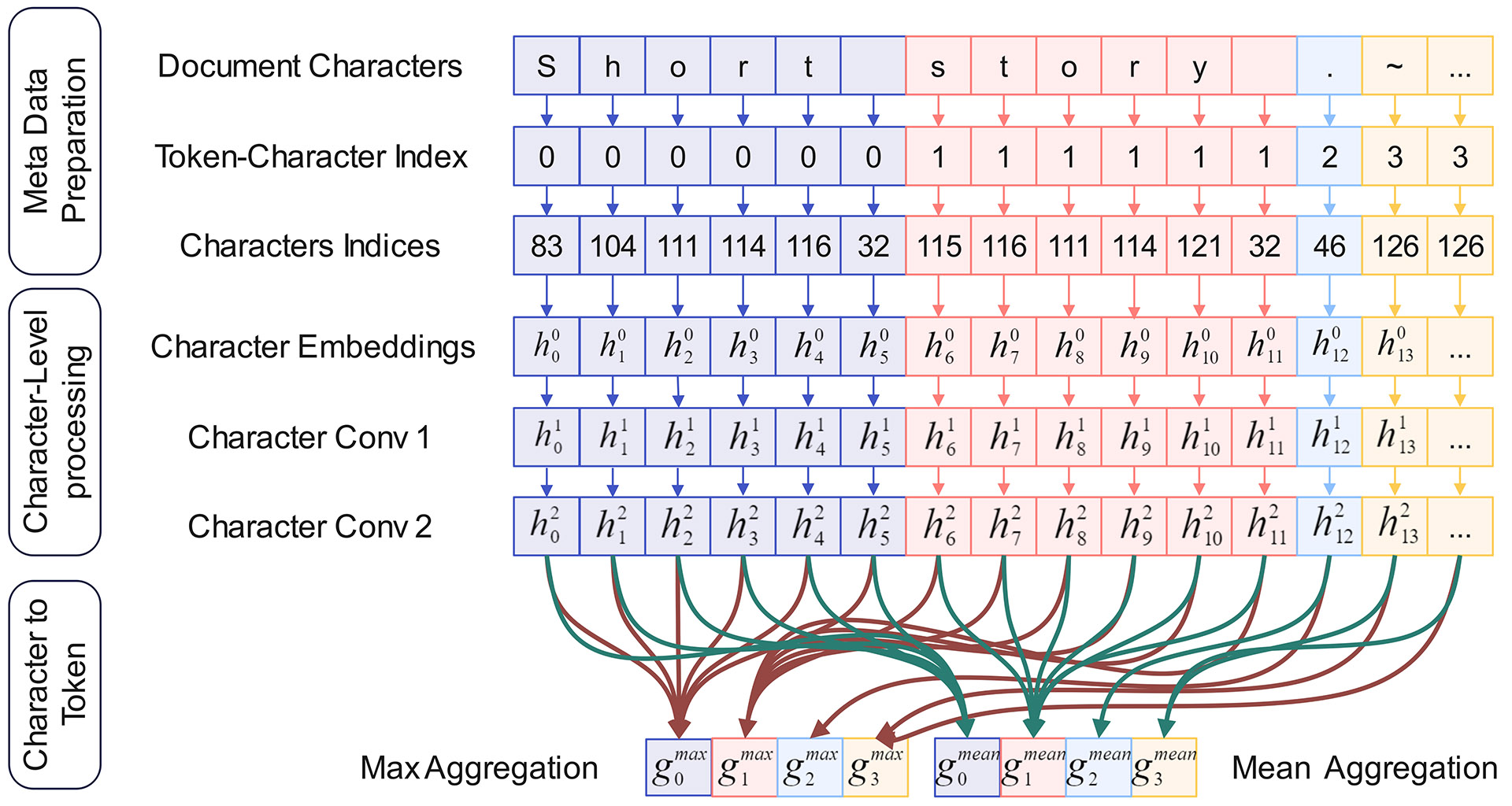}
  \caption{The aggregation of character-level features into token-level features \citep{rastakhiz2024quickcharnet}}
  \label{fig:character_feature_aggregation} 
\end{figure}

 \textbf{Character-to-Token Representation:}
The following equation describes the computation of the convolution operation.
\begin{equation} \label{eq_convolution}
  \mathbf{H}_i^{out}(C_{\text{out}})=\mathbf{b}_{C_{out}}+ \sum_{k = 0}^{C_{in} - 1} \mathbf{W}_{C_{\text{out}}}^k * \mathbf{H}_i^{in}(k)
\end{equation}
here, $\mathbf{H}_i^{out}(C_{\text{out}})$ represents the output for sample $i$ in the batch, and $C_{\text{out}}$ represents the number of output channels. $\mathbf{b}_{C{\text{out}}}$ is the bias term associated with the output channels. Based on \citep{o2015introduction}, the output size is computed through the following formula:
\begin{equation} \label{eq_convolution_out_size}
  n_{\text {out }}=\left\lfloor\frac{n_{\text {in }}+2 p-k}{s}\right\rfloor+1
\end{equation}
where $n_{\text {in}}$ and $n_{\text {out}}$ are the input and output sequence lengths, respectively; $k$ is the kernel size, $p$ is the padding size, and $s$ is the stride.
Character feature aggregation into token-level representation is illustrated in Figure \ref{fig:character_feature_aggregation}.

\par \textbf{Generate graph:} This component of the model, due to its close relation to data processing, has been described in Section \ref{section:graph_data_generation}; however, it is also considered a part of the model.

\par \textbf{CNN-GNN Combination Layer:}

\par
\begin{figure}[!htbp]
  \centering
  \includegraphics[width=\columnwidth]{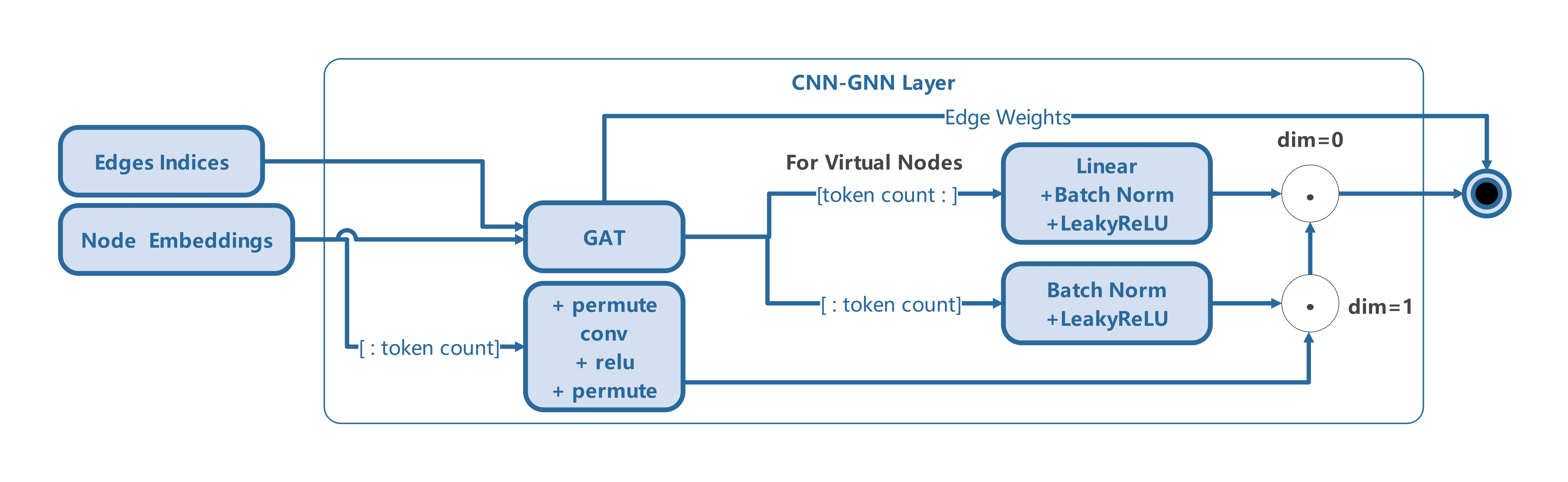}
  \caption{The CNN-GNN layer consists of a GAT/Attention layer and a 1D convolutional layer.}
  \label{fig:cnn_gnn_layer}
\end{figure}

 This layer is a combination of a convolution layer and graph neural network; hence, it is referred to as CNN-GNN. The GNN and CNN layers operate in parallel, as illustrated in Figure \ref{fig:cnn_gnn_layer}, to harness the strengths of both models and compensate for each other's limitations. For the GNN layer, either an improved version of the Graph Attention Network (GAT) \citep{brody2022attentive} or a modified version of a sparse attention layer is used. For the CNN layer, a one-dimensional convolution with kernel size three and padding one is employed.

\par Both CNNs and GNNs offer significant advantages for text processing. Notably, CNNs are well-suited for learning local spatial or temporal patterns, while GNNs are effective at capturing long-range, irregular, and complex dependencies. Additionally, both architectures support fast and parallel processing on various hardware platforms. However, aspects such as bias and variance warrant further research and consideration. The outputs of both layers are combined and passed to subsequent layers, allowing the model to efficiently leverage both local and global information.

\par The Graph Attention Network (GAT), introduced in \citep{veličković2018graph}, uses masked self-attention layers to assign different weights to different nodes, eliminating the need for costly matrix operations. Additionally, it does not require prior knowledge of the graph structure. However, as noted in \citep{brody2022attentive}, the original GAT has certain limitations, most notably, it computes only static attention, which assigns neighbor weights based solely on node scores. To address this, a dynamic GAT approach was proposed, offering improved accuracy and greater robustness to noise. The equations for the GAT layer are provided below:
\begin{equation} \label{eq_4}
  e(\mathbf{h}_i^{\ell},\mathbf{h}_j^{\ell}) = \mathbf{a}^{\top}\mathrm{LeakyReLU} \left(\mathbf{\Theta}[\mathbf{h}_i^{\ell} \, \Vert \, \mathbf{h}_j^{\ell}]
  \right)
\end{equation}
\begin{equation} \label{eq_5}
  \alpha_{i,j} = \frac{ \exp\left(e(\mathbf{h}_i^{\ell} ,\mathbf{h}_j^{\ell})\right)}{\sum_{k \in \mathcal{N}(i) \cup \{ j \}}\exp\left(e(\mathbf{h}_i^{\ell},\mathbf{h}_k^{\ell})\right)}
\end{equation}
where $\alpha_{i,j}$ represents the dynamic attention coefficient used in the aggregation of neighboring nodes and in the self-update mechanism, allowing the model to assign greater weight to the most relevant neighbors.
\begin{equation} \label{eq_6}
  f_{\theta}^{(\ell+1)}( \mathbf{h}^{\ell}_i, \mathbf{h}^{\ell}_j ) = \alpha_{i,i}\mathbf{\Theta}\mathbf{x}_{i} + \sum_{j \in \mathcal{N}(i)} \alpha_{i,j}\mathbf{\Theta}\mathbf{x}_{j}
\end{equation}

\par Here is the modified multi-head attention mechanism, which operates over a sparse graph of text using both edge and token weights. First, the inputs are prepared as follows:
\[
\begin{aligned}
& H = \text{heads count}, 
\space\space  N = \text{batch token count},
\space\space  d = \text{input features} \\
& M = \frac{d}{H}, 
\quad E = \text{number of edges},
\quad \mathbf{x} \;\in\;\mathbb{R}^{N\times d}, 
\space\space 
\bm{e}\;\in\;\mathbb{Z}^{2\times E}
\quad  \\
& \bm{W}^q,\space\bm{W}^k,\space\bm{W}^{\mathrm{agg}},\space\bm{W}^{\mathrm{upd}}\;\in\;\mathbb{R}^{M\times M}. 
\end{aligned}
\]

Reshaping and Permuting the Tensors:

\begin{equation} \label{eq:permute_x}
x'_{\,h,i,m} \gets x_{\,i,\;h\,M + m}
\quad
h=0,\dots,H-1;\;i=0,\dots,N-1;\;m=0,\dots,M-1
\end{equation}
\begin{equation} \label{eq:permute_e}
e'_{h,p,j} \gets \bm{e}_{\,p,\;h\,N' + j}
\quad \Bigl(p\in\{0,1\};\;j=0,\dots,N'-1\Bigr)
\end{equation}
where \(N' = E/H\) and \( p \) represents the edge's node 0 or 1. Gather the features of the nodes connected by each edge:
\begin{equation} \label{eq:edges_nodes_features}
S_{h,p,j,m} = x'_{\,h,\;e'_{h,p,j},\;m}
\quad \in\;\mathbb{R}^{H\times 2\times N'\times M}
\end{equation}
Query, Key, and Attention Scores:
\begin{equation} \label{eq:query}
Q_{h,j,m} = \sum_{p=0}^1 \sum_{\ell=0}^{M-1} S_{h,p,j,\ell}\,W^q_{h,\ell,m}
\end{equation}
\begin{equation} \label{eq:key}
K_{h,j,m} = \sum_{p=0}^1 \sum_{\ell=0}^{M-1} S_{h,p,j,\ell}\,W^k_{h,\ell,m}
\end{equation}
\begin{equation} \label{eq:attention}
a_{h,j} = \sum_{m=0}^{M-1} Q_{h,j,m}\,K_{h,j,m} \quad,\quad
\alpha_{h,j} = \frac{\exp(a_{h,j})}{\displaystyle\sum_{j'=0}^{N'-1}\exp(a_{h,j'})}\,\frac{1}{\sqrt{M}}
\end{equation}
Value Update and Aggregation:

\begin{equation} \label{eq:value_update}
V^{\mathrm{upd}}_{h,i,m} = \sum_{\ell=0}^{M-1} x^{(1)}_{h,i,\ell}\,W^{\mathrm{upd}}_{h,\ell,m}
\end{equation}
\begin{equation} \label{eq:value_aggregate}
V^{\mathrm{agg}}_{h,j,m} = \sum_{p=0}^1 \sum_{\ell=0}^{M-1} S_{h,p,j,\ell}\,W^{\mathrm{agg}}_{h,\ell,m}
\end{equation}
\begin{equation} \label{eq:attention_output}
Y_{h,i,m} = V^{\mathrm{upd}}_{h,i,m} \;+\;
\sum_{j:\,e_{h,0,j}=i} \alpha_{h,j}\,V^{\mathrm{agg}}_{h,j,m} \;+\;
\sum_{j:\,e_{h,1,j}=i} \alpha_{h,j}\,V^{\mathrm{agg}}_{h,j,m}
\end{equation}

\begin{figure}[!ht]
  \centering 
  \includegraphics[width=\textwidth/2]{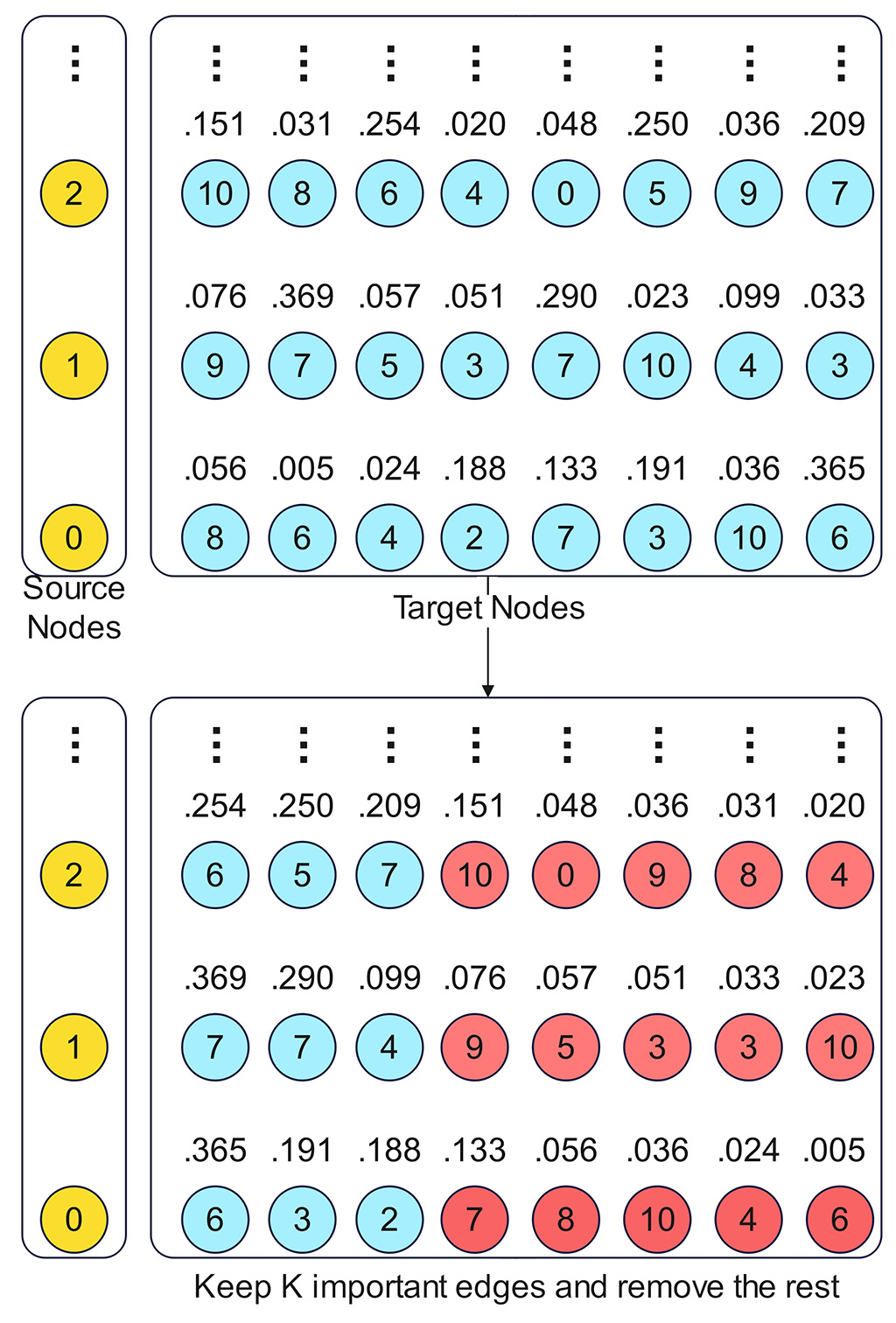}
  \caption{Retaining the \(k\) most important edges of the graph}
  \label{fig:keep_k_edges} 
\end{figure}

\textbf{Update graph:} After the first CNN-GNN layer, the attention weights from each layer are used to retain only the top few links for each node. The remaining edges are replaced with new ones using the graph generation approach described in Section \ref{section:graph_data_generation}. Figure \ref{fig:keep_k_edges} also illustrates this process.

\textbf{Positional encoding:} Positional encoding was first introduced in \citep{gehring2017convolutional} to provide a sense of order to token representations in convolutional networks. Since then, it has become a widely used technique in sequence-to-sequence models, particularly in attention mechanisms such as transformers. Unlike recurrent neural networks, transformers lack inherent sequential information due to their self-attention mechanism. Positional encoding addresses this limitation by injecting positional information directly into the input embeddings.
\par Typically, positional encoding involves creating fixed-length vectors that represent the position of tokens within the input sequence. By incorporating positional encoding, models can more effectively capture the order and relative positions of tokens, thereby enhancing the model's ability to understand sequential relationships between elements. The following formula is used here to create the positional encoding vectors:

\begin{equation} \label{eq:positional_encoding}
  \text{PE}(pos, 2i) = \sin\left(\frac{pos}{10000^{2i/d_{\text{model}}}}\right) + \cos\left(\frac{pos}{10000^{2i/d_{\text{model}}}}\right)
\end{equation}

\textbf{Token Embedding Injection:} The preparation of token embeddings is described in Section \ref{section:data_processing}. After the second graph construction layer, the model concatenates the generated representations with the corresponding token embeddings.

\par \textbf{Global Pooling:} This aggregates the token embeddings of each document in the batch and uses the ELU activation function, which helps address the vanishing gradient problem and produces a mean output closer to zero—similar to batch normalization, but with faster performance \citep{clevert2016fast}.

\par \textbf{MLP:} The following are the computation steps for the classification layers:
\begin{equation} \label{eq:mlp}
  \mathbf{g}( \mathbf{h} ) =
  \mathrm{ReLU} \left(\mathbf{h} \mathbf{\Theta}_1^T + \mathbf{b}_1
  \right), \\
  \mathbf{out}( \mathbf{g} ) = \mathbf{g} \mathbf{\Theta}_2^T + \mathbf{b}_2
\end{equation}
\par In the above equations, $\mathbf{h}$ represents the input to the MLP. The term $\mathbf{h} \mathbf{\Theta}_1^T + \mathbf{b}_1$ denotes the linear transformation applied to $\mathbf{h}$, where $\mathbf{\Theta}_1$ represents the weights and $\mathbf{b}_1$ the biases.

\par In the second equation, the output of first equation $\mathbf{g}$ is further processed by another linear transformation. The result of this transformation is the final output of the MLP.

\subsubsection{Memory and Computational Complexity of the Model}
\par The calculation of memory and computational complexity is as follows:

\begin{itemize}
\item One-dimensional convolution layers have complexity $O(nkd^2)$, where $n$ is the input length, $k$ is the kernel size, and $d$ is both the number of input features and filters \citep{vaswani2017attention}. \citet{KIRANYAZ2021107398} mention it as $O(nk)$; assuming fixed kernel size, embedding size, and number of filters, we can consider it $O(n)$.
\item As described in Section \ref{sec:preprocessing_complexity}, the graph generator layers have $O(n)$ time and space complexity relative to the input size. Relative to all variables, the complexity is $O(n(k_1 + k_2))$, where $k_1$ and $k_2$ are the number of lattice and random edges, respectively.
\item For GATv2 layers, according to \citep{brody2022attentive}, the complexity is $O(|\nu|dd' + |\epsilon|d')$, where $|\nu| = n$ is the number of input tokens and $|\epsilon| = kn$, with $k$ being a fixed number of neighbors. $d$ is the size of input features, and $d'$ is the size of output features. Assuming fixed $d$ and $d'$, the complexity simplifies to $O(n)$.
\item The torch scatter operation involves a memory mapping and a reduce function (mean and max in our case), both having $O(n)$ complexity.
\item Linear layers have $O(bd^2)$ complexity, where $b$ is the batch size and $d$ is the embedding size. Given that $b \ll n$ and $d \ll n$, this complexity can be considered negligible relative to $n$.
\item All activation functions used in the model have $O(n)$ complexity.
\item The embedding layer and embedding injection steps also have $O(n)$ complexity.
\end{itemize}

\par Therefore, the model has overall $O(n)$ time and space complexity relative to the input length $n$. However, beyond input length, some layers exhibit higher complexity relative to embedding size, number of features, number of filters $d$, and batch size $b$. Taking all variables into account, the overall complexity is:
\[
 O(n(d^2+(k_1+k_2).d)+b.d)
\]
or, in simplified form: 
\begin{equation} \label{eq:model_complexity}
 O(n(d^2+(k_1+k_2).d))
\end{equation}

\subsection{Datasets}
\label{section:datasets}

\par In this project, we aimed to use standard datasets with various features for the tasks of topic classification
and sentiment analysis. The datasets listed in Table \ref{tab:datasets} were used for the target tasks:

\begin{table*}[ht] 
        \caption{Information on the datasets used in this paper}
        \label{tab:datasets}
    \centering
    \resizebox{\textwidth}{!}{%
        \begin{tabular}{|p{4.8cm}|r|r|c|c|}
            \hline
            \textbf{Dataset}           & \textbf{Samples} & \textbf{Split}   & \textbf{Classes} & \textbf{Classes Distribution} \\
            \hline
            IMDB \citep{maas2011Learning}             & 50{,}000            & 0.5              & 2                & 25{,}000 / 25000                 \\
            Movie Review (RT-2K) \citep{pang2004asentiment}      & 2{,}000             & 0.1 (10-fold-CV) & 2                & 1{,}000 / 1000                   \\
            AG-News \citep{antonio2004agnews}          & 127{,}600            & 0.06             & 4                & 31{,}900 / 31{,}900 / 31{,}900 / 31{,}900 \\
            Yelp \citep{zhang2015character}            & 35{,}626            & 0.1              & 12               & Descibed in the paper         \\
            Amazon-Review \citep{mcauley2013hidden,zhang2015character}   & 400{,}000            & 0.1              & 12               & Descibed in the paper         \\

            \hline
        \end{tabular}
    }
\end{table*}

\par \textbf{Sentiment Analysis:}
\par IMDB, from \citep{maas2011Learning}, contains 50{,}000 samples, with 25{,}000 for training and 25{,}000 for validation. This dataset is balanced between two classes of negative and positive samples.
\par MR (Movie Reviews) polarity, from \citep{pang2004asentiment}, contains 2{,}000 balanced polarity reviews from 312 authors. Like in the paper \citep{ionescu2019vector}, 10-fold cross-validation is also used to evaluate this
dataset.
\par A subset of Amazon-Review Polarity, from \citep{mcauley2013hidden,zhang2015character}, contains 4{,}000{,}000
samples, with 3{,}600{,}000 train samples and 400{,}000 test samples. Here, only the first 10\% of the data, including
360{,}000 train and 40{,}000 test samples, were used. The dataset contains balanced negative and positive samples.
\par Yelp Polarity, from \citep{zhang2015character}, contains 598{,}000 samples, with 560{,}000 train samples and 38{,}000 test samples. This dataset is also balanced with negative and positive sample classes. The proposed model was
trained five times on a 40\% random subset of this dataset.

\par \textbf{Topic Classification:}
\par AG-News subset, from \citep{zhang2015character}, is a subset of the corpus of more than 1 million news articles
from \citep{antonio2004agnews}. The dataset contains four categories, including `World', `Sports', `Business', and
`Sci/Tech`. In this dataset, each class has 30{,}000 training samples and 19{,}000 test samples.

\subsection{Ablation Study}
\label{section:ablation}
 
\par This section evaluates the model's performance across various configurations on standard datasets, such as the Rotten-Tomato Movie-Reviews (RT-2K) dataset from \citep{pang2004asentiment}. Key aspects analyzed include normalization methods, convolution types, embedding injection techniques, semantic injection strategies, positional encoding schemes, and the handling of stop words and punctuation.  
 
\subsubsection{Normalization Method}
 
\par Normalization plays a crucial role in standardizing features, improving learning speed, preventing gradient explosion or vanishing, and enhancing the generalizability of the model \citep{huang2023normalization}. In the current architecture, batch normalization is difficult to apply due to the varying lengths of input sequences. As a result, layer normalization is a more suitable alternative. However, when samples are vertically stacked, a customized implementation of layer normalization is required. Another viable approach is token-level normalization, which aligns more naturally with the model's data structure and offers faster computation. Experimental results also indicate that this method yields better overall performance.

\par Another normalization approach involves normalizing each feature across all tokens, which is also computationally efficient. Several normalization techniques were evaluated, and Table \ref{tab:best_normalization} presents the average of the best results obtained for each method. For each normalization type, the model was trained 30 times on the Rotten-Tomato Movie-Reviews (RT-2K) dataset from \citep{pang2004asentiment}. The best results from each configuration were averaged and summarized in Table \ref{tab:best_normalization}. As shown, normalizing each feature across all tokens achieves the best performance across most evaluation metrics, except for the loss, where it performs the worst. This method applies the following formula to each feature across all tokens after the GNN layer:

\begin{equation} \label{eq:normalization}
     y = \frac{\mathbf{x} - \mu}{ \sqrt{\sigma}} * \gamma + \beta
\end{equation}
where $x$ denotes the feature column, $\mu$ represents the mean of the feature column, $\sigma$ is the standard deviation, and $\gamma$ and $\beta$ are learnable parameters representing the scale (variance) and shift (bias) for the feature, respectively. 
 
\begin{table}[h]
    \caption{Different normalizations for the new architecture}
    \label{tab:best_normalization}
    \centering
    \resizebox{\columnwidth}{!}{%
        \begin{tabular}{|@{\makebox[2em][c]{\rownumber\space}}|l|c|c|c|c|c|c|}
            \hline
            \gdef\rownumber{\stepcounter{magicrownumbers}\arabic{magicrownumbers}}
            \textbf{Model}                              & \textbf{Loss}   & \textbf{Acc.}  & \textbf{Prec.} & \textbf{Recall} & \textbf{F1-Score} & \textbf{FLOPs(M)} \\
            \hline
            Without normalization                       & 0.7776          & 67.26          & 67.64          & 66.94           & 67.28             & 20.723            \\
            Cutom Layer Normalization                   & 0.8858          & 63.52          & 63.99          & 63.45           & 63.72             & 20.723            \\
            Normalization on each feature of all tokens & 0.9398          & \textbf{82.61} & \textbf{82.95} & \textbf{82.60}  & \textbf{82.78}    & 20.723            \\
            Normalization at token level                & \textbf{0.7611} & 69.85          & 70.21          & 69.81           & 70.01             & 20.723            \\
            \hline
        \end{tabular}%
    }
\end{table}

\subsubsection{Depth-wise Separable Convolutions}

\par As noted in several studies, depth-wise separable convolutions offer high computational efficiency with minimal performance degradation. We conducted a comparison between depth-wise separable convolutions and standard 1D convolutions. The results are summarized in Table \ref{tab:best_convolution}. As shown, depth-wise separable convolutions require significantly fewer FLOPs, often by an order of magnitude, compared to standard convolutions. However, the drop in performance was substantial, leading us to retain the standard 1D convolution in subsequent experiments. 

\begin{table}[h]
    \caption{Different convolution layers for the new architecture}
    \label{tab:best_convolution}
    \centering
    \resizebox{\columnwidth}{!}{%
        \begin{tabular}{|@{\makebox[2em][c]{\rownumber\space}}|l|c|c|c|c|c|c|c|}
            \hline
            \gdef\rownumber{\stepcounter{magicrownumbers}\arabic{magicrownumbers}}
            \textbf{Model}               & \textbf{Loss} & \textbf{Acc.}  & \textbf{Prec.} & \textbf{Recall} & \textbf{F1-Score} & \textbf{FLOPs(M)} & \textbf{Params(K)} \\
            \hline
            Basic Form of 1D Convolution & 0.9619        & \textbf{81.96} & \textbf{82.31} & \textbf{81.96}  & \textbf{82.13}    & 20.723            & 409                \\
            Depth-wise convolution 1     & 0.7153        & 76.38          & 77.43          & 76.17           & 71.53             & 6.401             & 344                \\
            \hline
        \end{tabular}%
    }
\end{table}

\subsubsection{Tokenization and Inject embeddings}
\par In this study, embeddings extracted from SpaCy, BERT, and GPT are compressed using UMAP and subsequently evaluated. These dimensionality-reduced embeddings are injected into the model to conduct performance comparisons and identify the most effective configuration for the proposed architecture.

\par
\begin{figure*}[!htb]
    \centering
    \includegraphics[width=\textwidth]{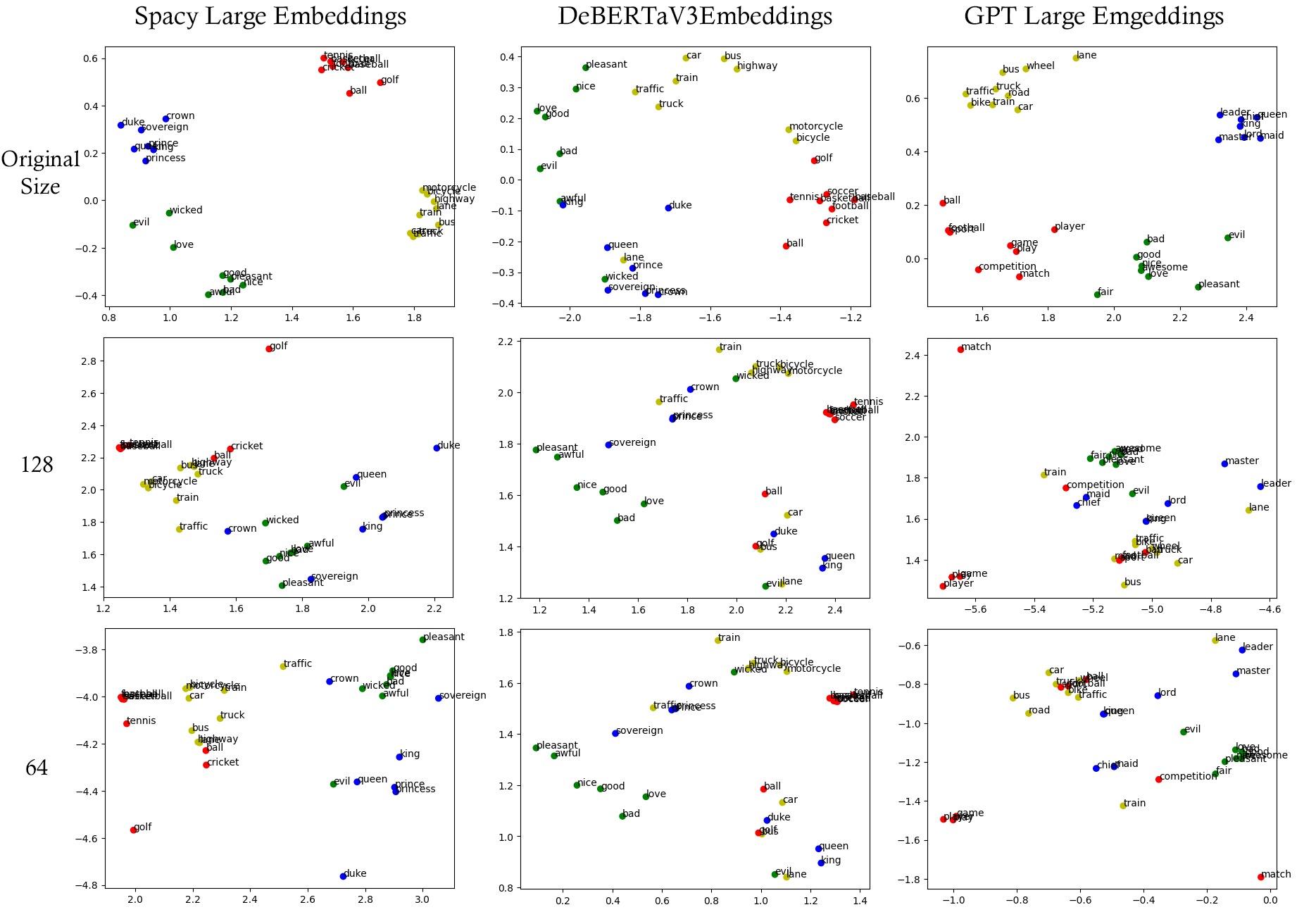}
    \caption{Comparison of original and size-reduced embeddings from three state-of-the-art sources using t-SNE visualization}
    \label{fig:reduced_embeddings_tsnes} 
\end{figure*}

 Figure \ref{fig:reduced_embeddings_tsnes} presents the t-SNE (t-distributed Stochastic Neighbor Embedding) visualizations of the original embeddings as well as the 64-dimensional and 128-dimensional reduced embeddings from the three model types (SpaCy, BERT, and GPT). As shown, all models exhibit relatively well-structured distributions in their original embedding spaces. Notably, GPT shows the most distinct separation of categories, with semantically related concepts tightly clustered within each class. DeBERTa demonstrates slightly more overlap between categories and concepts, though the structure remains largely preserved. 
\par For the dimensionality-reduced embeddings, all three models show some degree of distortion and category overlap due to information loss during compression. This effect is most visible in GPT, likely due to its higher initial embedding dimensionality. After reduction, SpaCy embeddings demonstrate the best category separation with the least overlap. DeBERTa shows marginally more overlap than SpaCy but still maintains a reasonably clear distinction between classes. 

\par \textbf{Token Sematic Embedding: }

\par Table \ref{tab:test_token_embeddings} presents the results of various experiments conducted on the RT-2K dataset using different embedding injection strategies. All forms of embedding injection improve performance over the base model without injection. Among them, DeBERTa achieves the highest accuracy, while SpaCy offers the fastest inference speed. The differences in computational efficiency can be attributed to the additional overhead introduced by embedding computation, as well as variations in tokenization strategies. Specifically, the SpaCy tokenizer generates the fewest tokens, whereas the Tiktoken tokenizer produces the most. The superior performance of DeBERTa may be attributed to its tokenizer design and the nature of the data it was pre-trained on.

\begin{table}[h]
    \caption{Token embeddings from different sources}
    \label{tab:test_token_embeddings}
    \centering
    \resizebox{\columnwidth}{!}{%
        \begin{tabular}{|@{\makebox[2em][c]{\rownumber\space}}|l|c|c|c|c|c|c|c|}
            \hline
            \gdef\rownumber{\stepcounter{magicrownumbers}\arabic{magicrownumbers}}
            \textbf{Model}              & \textbf{Loss}   & \textbf{Acc.}  & \textbf{Prec.} & \textbf{Recall} & \textbf{F1-Score} & \textbf{FLOPs(M)} & \textbf{Params(K)} \\
            \hline
            Without semantic embedding  & 0.9619          & 81.96 & 82.31 & 81.96  & 82.13    & 20.723            & 409                \\
            Inject SpaCy embeddings     & 0.9319          & 82.30          & 82.59          & 82.31           & 82.44             & 24.523            & 466                \\
            Inject debertav3 64-d embeddings & \textbf{0.7292} & \textbf{84.04} & \textbf{84.31} & \textbf{84.02}  & \textbf{84.16}    & 27.858            & 466                \\
            Inject gpt-large embeddings & 0.9180          & 82.52          & 82.96          & 82.46           & 82.71             & 27.941            & 553                \\
            \hline
        \end{tabular}%
    }
\end{table}

\par \textbf{Token Sentiment Embedding: }
This section examines whether incorporating the sentiment polarity of individual tokens can enhance the model's understanding of the overall document sentiment. However, since the sentiment of a token can be influenced by the broader document context, it is crucial to determine the optimal point of injection so that the model can effectively consider both local token-level and global document-level sentiment information. To this end, three distinct injection positions—along with their combinations—were evaluated. Each configuration was tested five times, and the average results are reported in Table \ref{tab:find_best_polarity_injection_placement}.

\begin{table}[h]
    \caption{Testing different positions and methods for injecting token polarity and subjectivity information. \(P_i\) denotes position \(i\), etc. \(P_1\) adds two additional channels to the third convolution layer, while \(P_2\) and \(P_3\) introduce new MLP or Convolution layers.}
    \label{tab:find_best_polarity_injection_placement}
    \centering
    \resizebox{\columnwidth}{!}{%
        \begin{tabular}{|@{\makebox[2em][c]{\rownumber\space}}|l|c|c|c|c|c|c|}
            \hline
            \gdef\rownumber{\stepcounter{magicrownumbers}\arabic{magicrownumbers}}
            \textbf{Model} & \textbf{Loss}   & \textbf{Acc.}  & \textbf{Prec.} & \textbf{Recall} & \textbf{F1-Score} & \textbf{FLOPs(M)} \\
            \hline
            No Injection   & 0.8217          & 83.30          & 83.61          & 83.18           & 83.39             & \textbf{27858}    \\
            $P_1$             & 0.8241          & 84.36          & 84.86          & 84.21           & 84.53             & 27890             \\
            $P_2$             & 0.8176          & 85.15          & 85.50          & 85.15           & 85.32             & 28541             \\
            $P_3$             & 0.8831          & 82.38          & 82.74          & 82.39           & 82.56             & 28541             \\
            $P_1$ and $P_2$      & 0.8265          & 85.00          & 85.18          & 84.93           & 85.06             & 28573             \\
            $P_1$ and $P_3$      & 0.8355          & 84.18          & 84.61          & 84.16           & 84.38             & 28573             \\
            $P_2$ and $P_3$      & 0.7696          & \textbf{85.78} & \textbf{86.06} & \textbf{85.78}  & \textbf{85.92}    & 29224             \\
            $P_1$, $P_2$ and $P_3$  & \textbf{0.6902} & 85.53          & 85.74          & 85.47           & 85.61             & 29256             \\
            \hline
        \end{tabular}%
    }
\end{table}

\begin{figure}[!htb]
    \centering
    \includegraphics[width=\columnwidth]{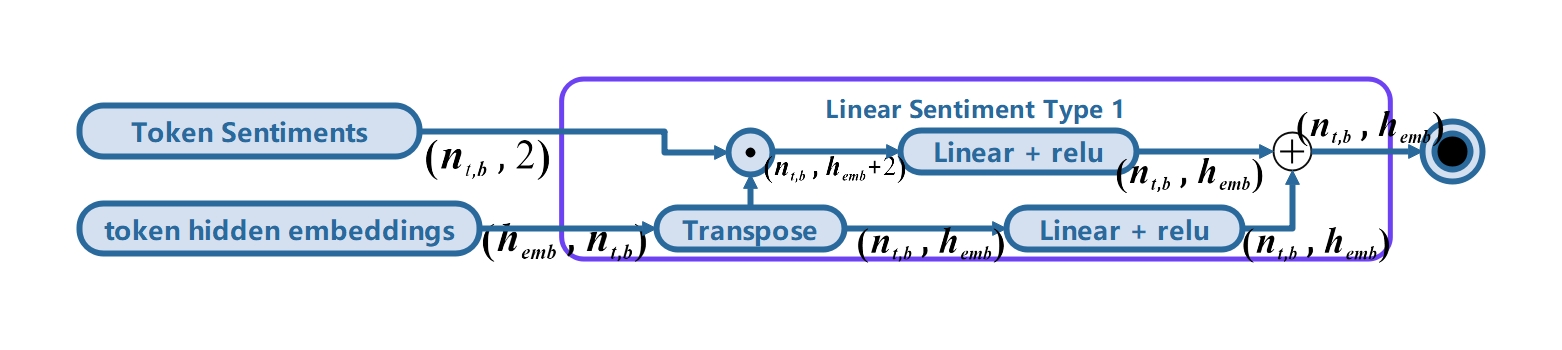}
    \caption{Architecture of the sentiment type 1 layer. In this layer, the two values for polarity and subjectivity are first concatenated with the token hidden embedding, then processed through linear layers.}
    \label{fig:sentiment_type_1} 
\end{figure}

\begin{figure}[!htb]
    \centering
    \includegraphics[width=\columnwidth]{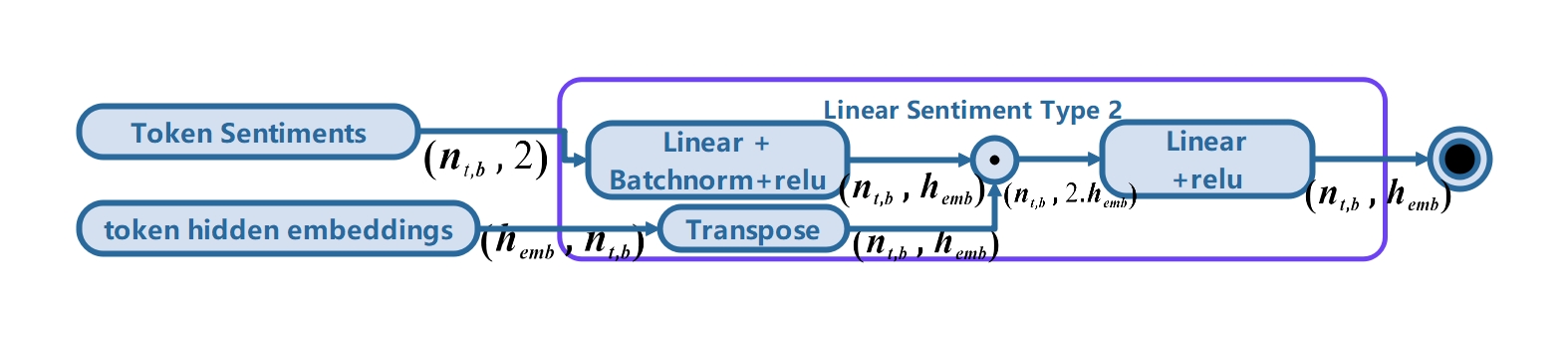}
    \caption{Architecture of the sentiment type 2 layer. In this layer, the polarity and subjectivity values are first processed by a linear layer for feature expansion, concatenated with the token hidden embedding, and then processed through another linear layer.}
    \label{fig:sentiment_type_2} 
\end{figure}

\begin{figure}[!htb]
    \centering
    \includegraphics[width=\columnwidth]{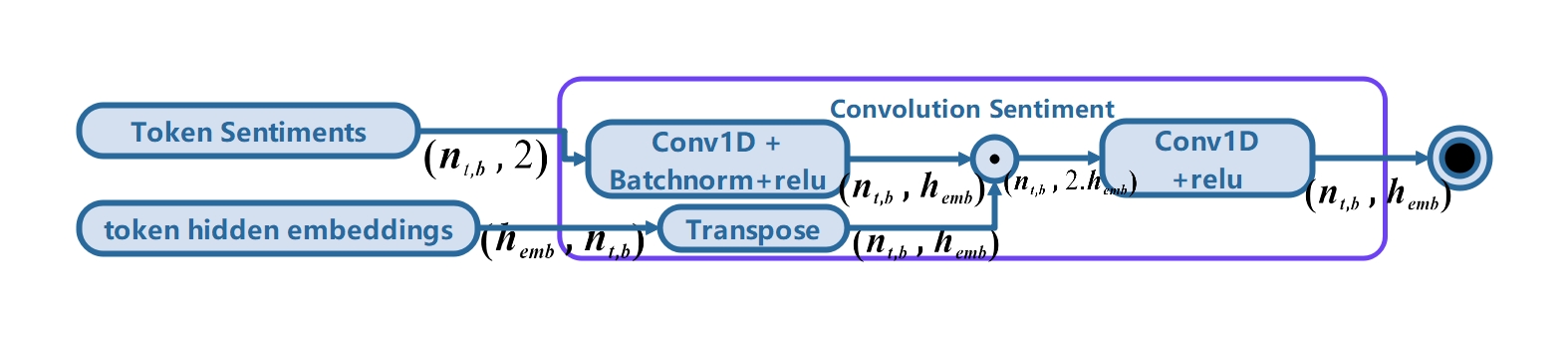}
    \caption{Architecture of the sentiment type 3 layer. This layer is similar to sentiment type 2 in Figure \ref{fig:sentiment_type_2}, except that it uses convolutional layers instead of linear layers.}
    \label{fig:sentiment_type_3} 
\end{figure}

\par Having identified the optimal position for injecting sentiment information, we next compare three types of sentiment processing layers, two MLP-based architectures and one convolutional layers, to determine which yields the best performance. The architectures are illustrated in Figures \ref{fig:sentiment_type_1}, \ref{fig:sentiment_type_2}, and \ref{fig:sentiment_type_3}. As shown in Table \ref{tab:find_best_polarity_injection_method}, the convolutional approach achieves the highest accuracy, albeit with a moderate increase in computational cost (FLOPs).   

\par Token-level sentiment is often influenced by neighboring tokens, such as contextual cues, phrases, or modifying words. Convolutional layers are particularly well-suited to capturing such local patterns, which likely contributes to their superior performance in this task. 

\begin{table}[h]
    \caption{Testing two custom sentiment injection layers: one using MLP and the other using convolutional layers. This test includes token semantic embeddings.}
    \label{tab:find_best_polarity_injection_method}
    \centering
    \resizebox{\columnwidth}{!}{%
        \begin{tabular}{|@{\makebox[2em][c]{\rownumber\space}}|l|c|c|c|c|c|c|c|}
            \hline
            \gdef\rownumber{\stepcounter{magicrownumbers}\arabic{magicrownumbers}}
            \textbf{Model}           & \textbf{Loss}   & \textbf{Acc.}  & \textbf{Prec.} & \textbf{Recall} & \textbf{F1-Score} & \textbf{FLOPs(M)} & \textbf{Parameters(K)} \\
            \hline
            Without sentiment injection & \textbf{0.7292} & \textbf{84.04} & \textbf{84.31} & \textbf{84.02}  & \textbf{84.16}    & 27.858            & 466                \\
            $P_2$ and $P_3$ custom layer (linear Type 1) & 0.7696          & 85.78 & 86.06 & 85.78  & 85.92    & 29224             & 483                    \\
            $P_2$ and $P_3$ custom layer (linear Type 2) & 0.7506          & 85.24 & 85.54 & 85.30  & 85.42    & 29224             & 483                    \\
            $P_2$ and $P_3$ custom layer (conv) & \textbf{0.7205} & \textbf{86.71} & \textbf{86.89} & \textbf{86.68}  & \textbf{86.78}    & 31.956            & 516                    \\
            \hline
        \end{tabular}
    }
\end{table}

\subsubsection{Stop-Words and Punctuations}

\par A common challenge in text analysis is the presence of high-frequency tokens with low informational value, such as stop words and punctuation. One effective approach to addressing this issue is adapting the subsampling technique within the graph generation process, inspired by its use in reducing the impact of frequent words in language modeling \citep{mikolov2013distributed}. This adaptation allows for the mitigation of both stop words and punctuation at the graph construction stage. 

\par The subsampling method probabilistically reduces the occurrence of frequent tokens by removing them from the training data with a certain probability, while simultaneously assigning higher weights to the remaining, more informative edges. This results in a more balanced representation that emphasizes semantically meaningful tokens, thereby enhancing both training efficiency and model performance. 
\par The proposed subsampling strategy is detailed in Algorithms \ref{alg:token_frequency}, \ref{alg:linear_subsampling}, \ref{alg:sigmoid_subsampling}, and \ref{alg:clustering_edge}. 

\begin{figure*}[!htb]
    \centering
    \includegraphics[width=\textwidth]{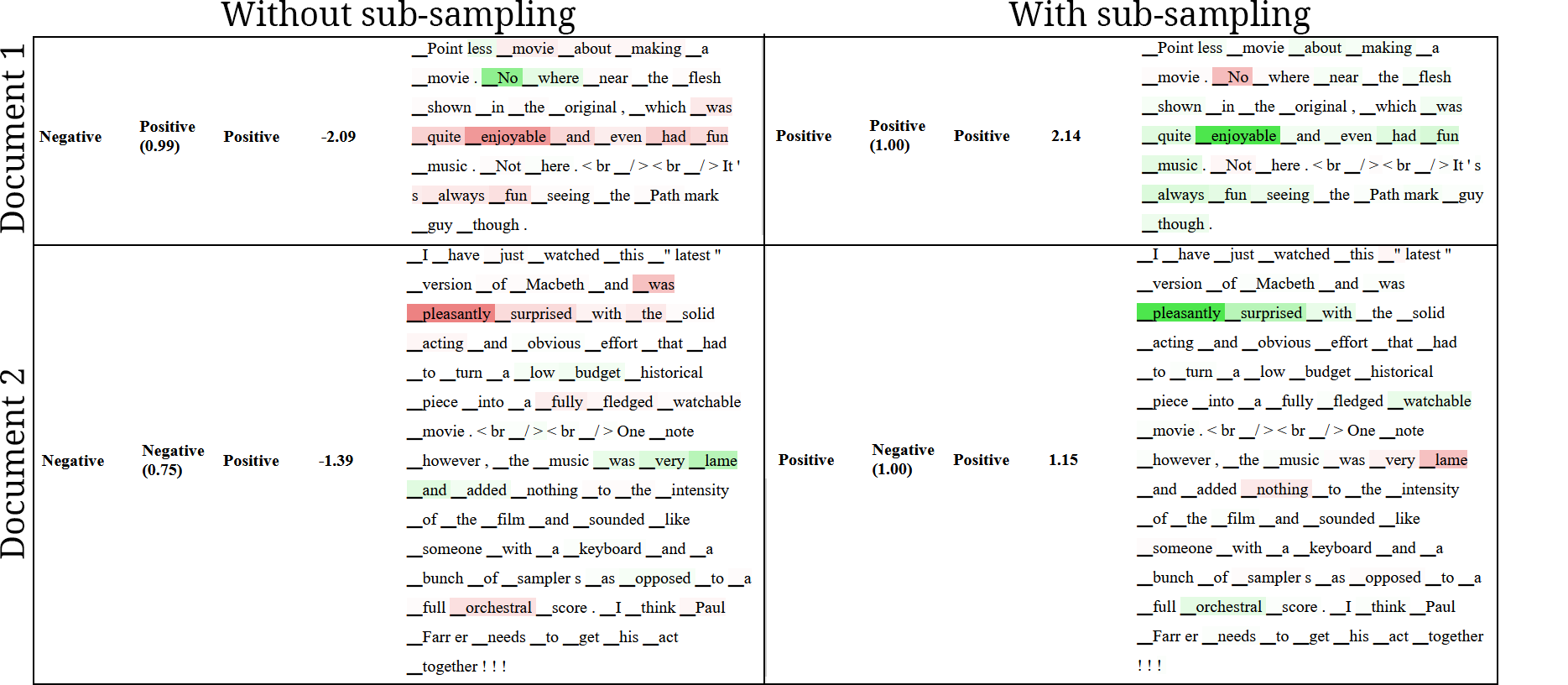}
    \caption{Comparison of layer attributions before and after implementing subsampling in graph generation}
    \label{fig:layers_attribution_1} 
\end{figure*}

\par Figure \ref{fig:layers_attribution_1} illustrates the attribution of after first token-level convolution layer in the model for two modes of with and without token frequencies sub-sampling for two documents. Documnet 1 shows that without subsampling it gave the wrong answer while with subsampling it gave correct one, and Document 2 shows the opposit. The attributions show the two models almost give attribution to the same tokens, while the model with subsampling has a little more tendency toward less frequent tokens. While expected that using subsampling improve the model, Table \ref{tab:subword_sampling} shows the opposit. The model without subsampling has slightly better result. Some reason that we examine in future works include: (1) BERT embedding can have some information about token imporance, punctuations, and stop words. (2) The generated graph has fixed lower bound number of edges per token, so the tokens with more frequency in document has more edges equivalent to term frequency. (3) The GNN uses attention mechanism, that can learn the weights for token embeddings, which can result in lower weight for less important tokens. (4) The integration of method into the model might need more in depth research and tests.

\par Figure \ref{fig:layers_attribution_1} visualizes the token attributions after the first token-level convolution layer for two versions of the model, one with and one without token frequency subsampling, on two example documents. Document 1 demonstrates that the model without subsampling produces an incorrect prediction, whereas the version with subsampling yields the correct result. In contrast, Document 2 shows the opposite behavior. 

\par The attribution maps indicate that both models largely attend to the same tokens; however, the model with subsampling exhibits a slightly stronger tendency to focus on less frequent tokens. While we expected subsampling to improve overall performance, the results in Table \ref{tab:subword_sampling} show the opposite: the model without subsampling achieves marginally better accuracy. 

Several potential explanations are considered for this outcome, which we plan to investigate in future work: 
\begin{enumerate}
    \item BERT embeddings may already encode information related to token importance, punctuation, and stop words, potentially reducing the added value of explicit subsampling.
    \item The graph generation process enforces a fixed lower bound on the number of edges per token, meaning that more frequent tokens naturally accumulate more connections, a behavior analogous to term frequency weighting.
    \item The GNN employs an attention mechanism that learns dynamic weights for token embeddings, potentially down-weighting unimportant tokens automatically.
    \item Further investigation is required to better integrate subsampling into the model architecture and training pipeline, including hyperparameter tuning and architectural adjustments.
\end{enumerate}
     
\begin{table}[h]
    \caption{Comparison of the model with and without subword sampling}
    \label{tab:subword_sampling}
    \centering
    \resizebox{\columnwidth}{!}{%
        \begin{tabular}{|@{\makebox[2em][c]{\rownumber\space}}|l|c|c|c|c|c|}
            \hline
            \gdef\rownumber{\stepcounter{magicrownumbers}\arabic{magicrownumbers}}
            \textbf{Model}           & \textbf{Loss}   & \textbf{Acc.}  & \textbf{Prec.} & \textbf{Recall} & \textbf{F1-Score}  \\
            \hline
            Without subsampling & 0.316          & \textbf{90.9} & \textbf{90.9} & \textbf{90.9}  & \textbf{90.9}    \\
            With raw subsampling & \textbf{0.250} & 90.89 & 90.89 & 90.89  & 90.89   \\
            \hline
        \end{tabular}%
    }
\end{table}

\subsection{Compare with Other Models}

\par In this section, the proposed model is compared with other state-of-the-art models across various datasets. Table \ref{tab:compare_model_compute} presents a list of the models along with their corresponding computational requirements. As shown, the proposed model has the lowest parameter count and requires fewer FLOPs compared to BERT and DistilBERT, making it particularly well-suited for deployment in resource-constrained environments. To benchmark the model, we use the hyperparameters listed in Table \ref{tab:model_hyper_parameters}.

\begin{table}[!h]
    \caption{Computational requirements of each model calculated on the AG-News dataset. Since the proposed model does not have a fixed size, values are averaged over a batch of 224 documents.}
    \label{tab:compare_model_compute}
    \centering
    \resizebox{\columnwidth}{!}{%
        \begin{tabular}{|@{\makebox[2em][c]{\rownumber\space}}|l|r|r|r|r|}
            \hline
            \gdef\rownumber{\stepcounter{magicrownumbers}\arabic{magicrownumbers}}
            \textbf{Model}       & \textbf{Batch Size} & \textbf{Hidden Dim}  & \textbf{FLOPs(M)} & \textbf{Parameters(M)} \\
            \hline
            BERT-Pretrained  \citep{devlin2019bert}       & 1    & 512    & 43536          & 109               \\
            DistilBERT-Pretrained  \citep{sanh2019distilbert} & 1    & 512    & 21769         & 67               \\ 
            Bi-GRU               & 1    & 256   & 60         & 33               \\
            proposed model        & 1    & 64   & 40         & 1.3               \\
            \hline
        \end{tabular}%
    }
\end{table}

\begin{table}[!h]
    \caption{Hyperparameter settings used to train the proposed model}
    \label{tab:model_hyper_parameters}
    \centering
    \resizebox{\columnwidth}{!}{%
        \begin{tabular}{|@{\makebox[2em][c]{\rownumber\space}}|l|l|}
            \hline
            \gdef\rownumber{\stepcounter{magicrownumbers}\arabic{magicrownumbers}}
            \textbf{Parameter}                 & \textbf{Value}                               \\
            \hline
            UTF-8 Character Set Size           & 12,288                                       \\
            Batch Size                         & 224 / 256 / 512                              \\
            Hidden Dimension                   & 64                                           \\
            Injected Embedding Dimension       & 64                                           \\
            Lattice Step Size                  & 2                                            \\
            Number of Lattice Edges            & 10                                           \\
            Number of Random Edges             & 6                                            \\
            Number of Epochs                   & 70                                           \\
            Dropout Rate                       & 0.2                                          \\
            Weight Decay                       & 0.000011                                     \\
            Initial Learning Rate              & 0.0032                                       \\
            Learning Rate Scheduler            & MultiStepLR                                  \\
            Learning Rate Milestones           & [15, 20, 30, 38, 40, 45, 50]                 \\
            LR Scheduler Decay Factor          & 0.5                                          \\
            Optimizer                          & AdamW                                        \\
            Loss Function                      & Binary / Categorical Cross-Entropy           \\
            \hline
        \end{tabular}%
    }
\end{table}

\par Tables \ref{tab:compare_model_on_ag_news}–\ref{tab:compare_model_on_amazon_reviews} compare the performance of the proposed model with that of other strong baselines on five standard text-classification datasets.

\begin{table}[!h]
    \caption{Performance on the AG-News dataset. Two of the models, BERT and DistilBERT, are pretrained.}
    \label{tab:compare_model_on_ag_news}
    \centering
    \resizebox{\columnwidth}{!}{%
        \begin{tabular}{|@{\makebox[2em][c]{\rownumber\space}}|l|r|r|r|r|r|}
            \hline
            \gdef\rownumber{\stepcounter{magicrownumbers}\arabic{magicrownumbers}}
            \textbf{Model}        & \textbf{Loss} & \textbf{Acc.} & \textbf{Prec.} & \textbf{Recall} & \textbf{F1-Score} \\
            \hline
            BERT-Pretrained       & 0.2435        & 94.49         & \textbf{94.89}          & \textbf{94.88}           & \textbf{94.88}             \\ 
            DistilBERT-Pretrained & 0.2064        & \textbf{94.79}         & 94.79          & 94.79           & 94.79             \\ 
            Naive Bayes             & -        & 89.58         & 88.89           & 88.89            & 88.89             \\ 
            Bi-GRU                  & -        & 91.53         & 91.53          & 91.52           & 91.53             \\ 
            proposed model        & \textbf{0.1045}        & \textit{93.08}         & \textit{93.07}          & \textit{93.08}           & \textit{93.07}             \\ 
            \hline
        \end{tabular}%
    }
\end{table}

\begin{table}[!h]
    \caption{Performance on the IMDB dataset}
    \label{tab:compare_model_on_IMDB_dataset}
    \centering
    \resizebox{\columnwidth}{!}{%
        \begin{tabular}{|@{\makebox[2em][c]{\rownumber\space}}|l|r|r|r|r|r|}
            \hline
            \gdef\rownumber{\stepcounter{magicrownumbers}\arabic{magicrownumbers}}
            \textbf{Model}         & \textbf{Loss} & \textbf{Acc.}  & \textbf{Prec.} & \textbf{Recall} & \textbf{F1-Score} \\
            \hline
            BERT-Pretrained        & 0.5272        & \textbf{92.08} & \textbf{92.10} & \textbf{92.08}  & \textbf{92.09}    \\ 
            DistilBERT-Pretrained  & 0.4209        & 91.83          & 91.83          & 91.83           & 91.83             \\ 
            Bi-GRU                 &  -             & 87.50         &  87.54        &  87.50         & 87.50           \\
            proposed model         & \textbf{0.2809} &  \textit{90.62}        &  \textit{90.62}          &  \textit{90.61}           &  \textit{90.62}             \\ 
            \hline
        \end{tabular}%
    }
\end{table}

\begin{table}[!h]
    \caption{Performance on the Rotten Tomatoes Movie Review (RT-2K) dataset}
    \label{tab:compare_model_on_MR_dataset}
    \centering
    \resizebox{\columnwidth}{!}{%
        \begin{tabular}{|@{\makebox[2em][c]{\rownumber\space}}|l|r|r|r|r|r|}
            \hline
            \gdef\rownumber{\stepcounter{magicrownumbers}\arabic{magicrownumbers}}
            \textbf{Model}        & \textbf{Loss} & \textbf{Acc.}  & \textbf{Prec.} & \textbf{Recall} & \textbf{F1-Score} \\
            \hline
            BERT-Pretrained       & \textbf{0.3308}        & 89.50          & 89.95          & 89.55           & 89.75             \\ 
            DistilBERT-Pretrained & 0.3410        & \textbf{90.00}          & \textbf{90.40}          & \textbf{90.00}           & \textbf{90.20}             \\ 
            Bi-GRU               &  -               & 77.85          & 78.21          & 77.72           & 77.66  \\
            proposed model        &  \textit{0.5940}        &  \textit{87.13} &  \textit{87.44} &  \textit{87.05}  &  \textit{87.24}    \\ 
            \hline
        \end{tabular}%
    }
\end{table}

\begin{table}[!h]
   \caption{Performance on a 40\% subset of the Yelp dataset}
    \label{tab:compare_model_on_yelp}
    \centering
    \resizebox{\columnwidth}{!}{%
        \begin{tabular}{|@{\makebox[2em][c]{\rownumber\space}}|l|r|r|r|r|r|}
            \hline
            \gdef\rownumber{\stepcounter{magicrownumbers}\arabic{magicrownumbers}}
            \textbf{Model}        & \textbf{Loss}   & \textbf{Acc.}  & \textbf{Prec.} & \textbf{Recall} & \textbf{F1-Score} \\
            \hline
            BERT-Pretrained       & 0.3275          & 95.86          & 95.86          & 95.85           & 95.86  \\ 
            DistilBERT-Pretrained & 0.2977          & \textbf{95.89}          & \textbf{95.89}          & \textbf{95.89}           & \textbf{95.89}  \\ 
            Proposed Model        & \textbf{0.1024} & \textit{95.61} & \textit{95.61} & \textit{95.61}  & \textit{95.61}    \\ 
            \hline
        \end{tabular}%
    }
\end{table}

\begin{table}[!h]
    \caption{Performance on a 10\% subset of the Amazon Reviews dataset}
    \label{tab:compare_model_on_amazon_reviews}
    \centering
    \resizebox{\columnwidth}{!}{%
        \begin{tabular}{|@{\makebox[2em][c]{\rownumber\space}}|l|r|r|r|r|r|}
            \hline
            \gdef\rownumber{\stepcounter{magicrownumbers}\arabic{magicrownumbers}}
            \textbf{Model}        & \textbf{Loss} & \textbf{Acc.} & \textbf{Prec.} & \textbf{Recall} & \textbf{F1-Score} \\
            \hline
            BERT-Pretrained       & 0.3865        & 94.26         & 94.26          & 94.26           & 94.26             \\
            DistilBERT-Pretrained & 0.1630        & 94.28         & 94.28          & 94.28           & 94.28             \\
            Proposed Model        & \textbf{0.1561}        & \textbf{94.37}         & \textbf{94.38}          & \textbf{94.37}          & \textbf{94.37}             \\
            \hline
        \end{tabular}%
    }
\end{table}

\paragraph{Analysis of Results:}
Overall, our proposed model achieves a dramatic reduction in compute (1.3 M parameters and 40 M FLOPs) compared to large pretrained transformers (up to 109 M parameters and 43 B FLOPs) while maintaining strong classification performance. On AG-News, it attains a competitive F1 of 93.07\% (Table \ref{tab:compare_model_on_ag_news}), trailing DistilBERT by only 1.71 percentage points (pp) despite being over 51x smaller. Similarly, on IMDB (Table \ref{tab:compare_model_on_IMDB_dataset}), our model yields 90.62\% F1, closing to pretrained and finetuned BERT (92.09\%) with less than 1.5 pp difference in accuracy.

However, on the small RT-2K sentiment benchmark (Table \ref{tab:compare_model_on_MR_dataset}), the proposed model underperforms pretrained baselines by about 2-3 pp, likely because it has fewer representational degrees of freedom to capture nuanced sentiment without pretraining. On large-scale subsets of Yelp and Amazon reviews (Tables \ref{tab:compare_model_on_yelp} and \ref{tab:compare_model_on_amazon_reviews}), our model remarkably matches or slightly exceeds DistilBERT (95.61\% vs. 95.89\% on Yelp, 94.37\% vs. 94.28\% on Amazon-Reviews), demonstrating strong scalability.

\paragraph{Discussion and Future Work: }
The results confirm that lightweight architectures can provide an excellent compute-accuracy trade-off for many text classification tasks, particularly when pretrained models are too expensive for edge or low-power deployment. The proposed model is highly flexible and offers multiple paths for further enhancement. For example, its architecture could be adapted to incorporate natural language principles, such as graph generation guided by linguistic rules. Potential future improvements include:

\begin{itemize}
    \item \textbf{Incorporating pretraining or transfer learning:} to close the remaining performance gap on smaller datasets such as RT-2K.
    \item \textbf{Applying knowledge distillation:} leveraging a larger teacher model to improve performance without increasing inference costs.
    \item \textbf{Exploring alternative mechanisms to regulate high-frequency tokens:} for instance, integrating token frequency constraints inspired by Zipf's law.
    \item \textbf{Investigating novel local information processing techniques:} exploring methods beyond traditional CNNs.
    \item \textbf{Analyzing the impact of architectural choices:} studying how different layer types affect the bias-variance trade-off.
    \item \textbf{Optimizing the architecture and preprocessing pipeline:} reducing unnecessary complexity and dependencies on metadata.
    \item \textbf{Extending the model to new tasks:} such as named entity recognition and text embedding generation for information retrieval systems.
    \item \textbf{Introducing dynamic token representations:} replacing fixed tokens with dynamic tokens through new layers and exploring alternative types of knowledge injection.
\end{itemize}

With these enhancements, the proposed approach has the potential to become an even more versatile and efficient solution across a wide range of natural language processing applications.

\section{Conclusion}
\label{conclusion}
\par This study introduced an efficient deep learning model designed to deliver high performance across texts of varying lengths. Ablation studies were conducted to assess the contribution of individual components, leading to the design of the final architecture. The proposed model was then rigorously evaluated on multiple text classification tasks and benchmarked against state-of-the-art models across diverse datasets. Experimental results demonstrate that the model achieves competitive accuracy while significantly reducing computational and memory requirements, making it a strong candidate for deployment in resource-constrained environments.

\bibliographystyle{unsrtnat}
\bibliography{references}  

\begin{thebibliography}{60}
\providecommand{\natexlab}[1]{#1}
\providecommand{\url}[1]{\texttt{#1}}
\expandafter\ifx\csname urlstyle\endcsname\relax
  \providecommand{\doi}[1]{doi: #1}\else
  \providecommand{\doi}{doi: \begingroup \urlstyle{rm}\Url}\fi

\bibitem[Rastakhiz(2025)]{rastakhiz2025gnn}
Fardin Rastakhiz.
\newblock {CNN-GNN-Text}, 2025.
\newblock URL \url{https://doi.org/10.17632/d3cw4gyz85.3}.

\bibitem[Paleyes et~al.(2022)Paleyes, Urma, and Lawrence]{paleyes2022challenges}
Andrei Paleyes, Raoul-Gabriel Urma, and Neil~D Lawrence.
\newblock Challenges in deploying machine learning: a survey of case studies.
\newblock \emph{ACM computing surveys}, 55\penalty0 (6):\penalty0 1--29, 2022.

\bibitem[Sharir et~al.(2020)Sharir, Peleg, and Shoham]{sharir2020cost}
Or~Sharir, Barak Peleg, and Yoav Shoham.
\newblock The cost of training nlp models: A concise overview.
\newblock \emph{arXiv preprint arXiv:2004.08900}, 2020.

\bibitem[Sevilla et~al.(2022)Sevilla, Heim, Ho, Besiroglu, Hobbhahn, and Villalobos]{sevilla2022compute}
Jaime Sevilla, Lennart Heim, Anson Ho, Tamay Besiroglu, Marius Hobbhahn, and Pablo Villalobos.
\newblock Compute trends across three eras of machine learning.
\newblock In \emph{2022 International Joint Conference on Neural Networks (IJCNN)}, pages 1--8. IEEE, 2022.

\bibitem[Strubell et~al.(2020)Strubell, Ganesh, and McCallum]{strubell2020energy}
Emma Strubell, Ananya Ganesh, and Andrew McCallum.
\newblock Energy and policy considerations for modern deep learning research.
\newblock In \emph{Proceedings of the AAAI conference on artificial intelligence}, volume~34, pages 13693--13696, 2020.

\bibitem[Benitez et~al.(2018)Benitez, Ray, and Henseler]{benitez2018impact}
Jose Benitez, Gautam Ray, and J{\"o}rg Henseler.
\newblock Impact of information technology infrastructure flexibility on mergers and acquisitions.
\newblock \emph{MIS quarterly}, 42\penalty0 (1):\penalty0 25--A12, 2018.

\bibitem[Amodei and Hernandez(2018)]{amodei2018ai}
Dario Amodei and Danny Hernandez.
\newblock Ai and compute.
\newblock \emph{OpenAI Blog https://openai.com/blog/ai-and-compute/}, 5 2018.

\bibitem[Han et~al.(2015)Han, Mao, and Dally]{han2015deep}
Song Han, Huizi Mao, and William~J Dally.
\newblock Deep compression: Compressing deep neural networks with pruning, trained quantization and huffman coding.
\newblock \emph{arXiv preprint arXiv:1510.00149}, 2015.

\bibitem[Hinton(2015)]{hinton2015distilling}
Geoffrey Hinton.
\newblock Distilling the knowledge in a neural network.
\newblock \emph{arXiv preprint arXiv:1503.02531}, 2015.

\bibitem[Li and Meng(2023)]{hengyi2023hardware}
Hengyi Li and Lin Meng.
\newblock Hardware-aware approach to deep neural network optimization.
\newblock \emph{Neurocomputing}, 559:\penalty0 126808, 2023.
\newblock ISSN 0925-2312.
\newblock \doi{https://doi.org/10.1016/j.neucom.2023.126808}.
\newblock URL \url{https://www.sciencedirect.com/science/article/pii/S0925231223009311}.

\bibitem[Khurana et~al.(2023)Khurana, Koli, Khatter, and Singh]{khurana2023natural}
Diksha Khurana, Aditya Koli, Kiran Khatter, and Sukhdev Singh.
\newblock Natural language processing: state of the art, current trends and challenges.
\newblock \emph{Multimedia tools and applications}, 82\penalty0 (3):\penalty0 3713--3744, 2023.

\bibitem[Reusens et~al.(2024)Reusens, Stevens, Tonglet, De~Smedt, Verbeke, Vanden~Broucke, and Baesens]{reusens2024evaluating}
Manon Reusens, Alexander Stevens, Jonathan Tonglet, Johannes De~Smedt, Wouter Verbeke, Seppe Vanden~Broucke, and Bart Baesens.
\newblock Evaluating text classification: A benchmark study.
\newblock \emph{Expert Systems with Applications}, page 124302, 2024.

\bibitem[Fields et~al.(2024)Fields, Chovanec, and Madiraju]{fields2024survey}
John Fields, Kevin Chovanec, and Praveen Madiraju.
\newblock A survey of text classification with transformers: How wide? how large? how long? how accurate? how expensive? how safe?
\newblock \emph{IEEE Access}, 2024.

\bibitem[Hu et~al.(2024)Hu, Hou, and Liu]{hu2024deep}
Zhentao Hu, Wei Hou, and Xianxing Liu.
\newblock Deep learning for named entity recognition: a survey.
\newblock \emph{Neural Computing and Applications}, 36\penalty0 (16):\penalty0 8995--9022, 2024.

\bibitem[Daneshfar et~al.(2024)Daneshfar, Soleymanbaigi, Nafisi, and Yamini]{daneshfar2024elastic}
Fatemeh Daneshfar, Sayvan Soleymanbaigi, Ali Nafisi, and Pedram Yamini.
\newblock Elastic deep autoencoder for text embedding clustering by an improved graph regularization.
\newblock \emph{Expert Systems with Applications}, 238:\penalty0 121780, 2024.

\bibitem[Li et~al.(2024{\natexlab{a}})Li, Tang, Zhao, Nie, and Wen]{li2024pre}
Junyi Li, Tianyi Tang, Wayne~Xin Zhao, Jian-Yun Nie, and Ji-Rong Wen.
\newblock Pre-trained language models for text generation: A survey.
\newblock \emph{ACM Computing Surveys}, 56\penalty0 (9):\penalty0 1--39, 2024{\natexlab{a}}.

\bibitem[He et~al.(2024)He, Liang, Jiao, Zhang, Yang, Wang, Tu, Shi, and Wang]{he2024exploring}
Zhiwei He, Tian Liang, Wenxiang Jiao, Zhuosheng Zhang, Yujiu Yang, Rui Wang, Zhaopeng Tu, Shuming Shi, and Xing Wang.
\newblock Exploring human-like translation strategy with large language models.
\newblock \emph{Transactions of the Association for Computational Linguistics}, 12:\penalty0 229--246, 2024.

\bibitem[Li et~al.(2024{\natexlab{b}})Li, Zhou, Huang, Cheng, and Chen]{li2024eliciting}
Jiahuan Li, Hao Zhou, Shujian Huang, Shanbo Cheng, and Jiajun Chen.
\newblock Eliciting the translation ability of large language models via multilingual finetuning with translation instructions.
\newblock \emph{Transactions of the Association for Computational Linguistics}, 12:\penalty0 576--592, 2024{\natexlab{b}}.

\bibitem[Maurya et~al.(2023)Maurya, Singh, and Maurya]{maurya2023deceptive}
Sushil~Kumar Maurya, Dinesh Singh, and Ashish~Kumar Maurya.
\newblock Deceptive opinion spam detection approaches: a literature survey.
\newblock \emph{Applied intelligence}, 53\penalty0 (2):\penalty0 2189--2234, 2023.

\bibitem[Yadav and Kaur(2023)]{yadav2023machine}
Deepak Yadav and Er~Bhupinder Kaur.
\newblock Machine learning models for email spam detection: A review.
\newblock In \emph{2023 Second International Conference On Smart Technologies For Smart Nation (SmartTechCon)}, pages 1109--1114. IEEE, 2023.

\bibitem[Young et~al.(2018)Young, Hazarika, Poria, and Cambria]{young2018recent}
Tom Young, Devamanyu Hazarika, Soujanya Poria, and Erik Cambria.
\newblock Recent trends in deep learning based natural language processing.
\newblock \emph{ieee Computational intelligenCe magazine}, 13\penalty0 (3):\penalty0 55--75, 2018.

\bibitem[Vaswani et~al.(2017)Vaswani, Shazeer, Parmar, Uszkoreit, Jones, Gomez, Kaiser, and Polosukhin]{vaswani2017attention}
Ashish Vaswani, Noam Shazeer, Niki Parmar, Jakob Uszkoreit, Llion Jones, Aidan~N Gomez, {\L}ukasz Kaiser, and Illia Polosukhin.
\newblock Attention is all you need.
\newblock \emph{Advances in neural information processing systems}, 30, 2017.

\bibitem[Brown et~al.(2020)Brown, Mann, Ryder, Subbiah, Kaplan, Dhariwal, Neelakantan, Shyam, Sastry, Askell, et~al.]{brown2020language}
Tom Brown, Benjamin Mann, Nick Ryder, Melanie Subbiah, Jared~D Kaplan, Prafulla Dhariwal, Arvind Neelakantan, Pranav Shyam, Girish Sastry, Amanda Askell, et~al.
\newblock Language models are few-shot learners.
\newblock \emph{Advances in neural information processing systems}, 33:\penalty0 1877--1901, 2020.

\bibitem[Liu et~al.(2020)Liu, Zhou, Zhao, Wang, Deng, and Ju]{liu2020fastbert}
Weijie Liu, Peng Zhou, Zhe Zhao, Zhiruo Wang, Haotang Deng, and Qi~Ju.
\newblock Fastbert: a self-distilling bert with adaptive inference time.
\newblock \emph{arXiv preprint arXiv:2004.02178}, 2020.

\bibitem[Rastakhiz et~al.(2024)Rastakhiz, Eftekhari, and Vahdati]{rastakhiz2024quickcharnet}
Fardin Rastakhiz, Mahdi Eftekhari, and Sahar Vahdati.
\newblock Quickcharnet: An efficient url classification framework for enhanced search engine optimization.
\newblock \emph{IEEE Access}, 12:\penalty0 156965--156979, 2024.
\newblock \doi{10.1109/ACCESS.2024.3484578}.

\bibitem[Romero et~al.(2022)Romero, Celard, Sorribes-Fdez, Vieira, Iglesias, and Borrajo]{romero2022mobydeep}
Rub{\'e}n Romero, Pedro Celard, Jos{\'e}~Manuel Sorribes-Fdez, A~Seara Vieira, Eva~Lorenzo Iglesias, and L~Borrajo.
\newblock Mobydeep: A lightweight cnn architecture to configure models for text classification.
\newblock \emph{Knowledge-Based Systems}, 257:\penalty0 109914, 2022.

\bibitem[Chen et~al.(2024)Chen, Li, Han, Ren, and Yang]{chen2024review}
Fanghui Chen, Shouliang Li, Jiale Han, Fengyuan Ren, and Zhen Yang.
\newblock Review of lightweight deep convolutional neural networks.
\newblock \emph{Archives of Computational Methods in Engineering}, 31\penalty0 (4):\penalty0 1915--1937, 2024.

\bibitem[Yang et~al.(2021)Yang, Hu, Shi, Ji, Li, and Nie]{yang2021hgat}
Tianchi Yang, Linmei Hu, Chuan Shi, Houye Ji, Xiaoli Li, and Liqiang Nie.
\newblock Hgat: Heterogeneous graph attention networks for semi-supervised short text classification.
\newblock \emph{ACM Transactions on Information Systems (TOIS)}, 39\penalty0 (3):\penalty0 1--29, 2021.

\bibitem[Shirzad et~al.(2023)Shirzad, Velingker, Venkatachalam, Sutherland, and Sinop]{shirzad2023exphormer}
Hamed Shirzad, Ameya Velingker, Balaji Venkatachalam, Danica~J Sutherland, and Ali~Kemal Sinop.
\newblock Exphormer: Sparse transformers for graphs.
\newblock In \emph{International Conference on Machine Learning}, pages 31613--31632. PMLR, 2023.

\bibitem[Jaszczur et~al.(2021)Jaszczur, Chowdhery, Mohiuddin, Kaiser, Gajewski, Michalewski, and Kanerva]{jaszczur2021sparse}
Sebastian Jaszczur, Aakanksha Chowdhery, Afroz Mohiuddin, Lukasz Kaiser, Wojciech Gajewski, Henryk Michalewski, and Jonni Kanerva.
\newblock Sparse is enough in scaling transformers.
\newblock \emph{Advances in Neural Information Processing Systems}, 34:\penalty0 9895--9907, 2021.

\bibitem[Zhou et~al.(2020)Zhou, Cui, Hu, Zhang, Yang, Liu, Wang, Li, and Sun]{zhou2020graph}
Jie Zhou, Ganqu Cui, Shengding Hu, Zhengyan Zhang, Cheng Yang, Zhiyuan Liu, Lifeng Wang, Changcheng Li, and Maosong Sun.
\newblock Graph neural networks: A review of methods and applications.
\newblock \emph{AI Open}, 1:\penalty0 57--81, 2020.
\newblock ISSN 2666-6510.
\newblock \doi{https://doi.org/10.1016/j.aiopen.2021.01.001}.
\newblock URL \url{https://www.sciencedirect.com/science/article/pii/S2666651021000012}.

\bibitem[Child et~al.(2019)Child, Gray, Radford, and Sutskever]{child2019generating}
Rewon Child, Scott Gray, Alec Radford, and Ilya Sutskever.
\newblock Generating long sequences with sparse transformers.
\newblock \emph{arXiv preprint arXiv:1904.10509}, 2019.

\bibitem[Tay et~al.(2022)Tay, Dehghani, Bahri, and Metzler]{tay2022efficient}
Yi~Tay, Mostafa Dehghani, Dara Bahri, and Donald Metzler.
\newblock Efficient transformers: A survey.
\newblock \emph{ACM Computing Surveys}, 55\penalty0 (6):\penalty0 1--28, 2022.

\bibitem[Ramp{\'a}{\v{s}}ek et~al.(2022)Ramp{\'a}{\v{s}}ek, Galkin, Dwivedi, Luu, Wolf, and Beaini]{rampavsek2022recipe}
Ladislav Ramp{\'a}{\v{s}}ek, Michael Galkin, Vijay~Prakash Dwivedi, Anh~Tuan Luu, Guy Wolf, and Dominique Beaini.
\newblock Recipe for a general, powerful, scalable graph transformer.
\newblock \emph{Advances in Neural Information Processing Systems}, 35:\penalty0 14501--14515, 2022.

\bibitem[Xie et~al.(2025)Xie, Dong, Yang, Luo, Ren, Zhang, He, Jia, Yang, Jiang, Gao, and Chen]{XIE2025112631}
Hui Xie, Zexiao Dong, Huiting Yang, Yanxia Luo, Shenghan Ren, Pengyuan Zhang, Jiangshan He, Chunli Jia, Yuqiang Yang, Mingzhe Jiang, Xinbo Gao, and Xueli Chen.
\newblock Cnn-transformer network for student learning effect prediction using eeg signals based on spatio-temporal feature fusion.
\newblock \emph{Applied Soft Computing}, 170:\penalty0 112631, 2025.
\newblock ISSN 1568-4946.
\newblock \doi{https://doi.org/10.1016/j.asoc.2024.112631}.
\newblock URL \url{https://www.sciencedirect.com/science/article/pii/S1568494624014054}.

\bibitem[He et~al.(2021{\natexlab{a}})He, Gao, and Chen]{he2021debertav3}
Pengcheng He, Jianfeng Gao, and Weizhu Chen.
\newblock Debertav3: Improving deberta using electra-style pre-training with gradient-disentangled embedding sharing, 2021{\natexlab{a}}.

\bibitem[Devlin et~al.(2019)Devlin, Chang, Lee, and Toutanova]{devlin2019bert}
Jacob Devlin, Ming-Wei Chang, Kenton Lee, and Kristina Toutanova.
\newblock Bert: Pre-training of deep bidirectional transformers for language understanding, 2019.
\newblock URL \url{https://arxiv.org/abs/1810.04805}.

\bibitem[Liu(2019)]{liu2019roberta}
Yinhan Liu.
\newblock Roberta: A robustly optimized bert pretraining approach.
\newblock \emph{arXiv preprint arXiv:1907.11692}, 364, 2019.

\bibitem[He et~al.(2021{\natexlab{b}})He, Liu, Gao, and Chen]{he2020deberta}
Pengcheng He, Xiaodong Liu, Jianfeng Gao, and Weizhu Chen.
\newblock Deberta: Decoding-enhanced bert with disentangled attention.
\newblock In \emph{International Conference on Learning Representations}, 2021{\natexlab{b}}.
\newblock URL \url{https://openreview.net/forum?id=XPZIaotutsD}.

\bibitem[Kaiser(2008)]{kaiser2008mean}
Marcus Kaiser.
\newblock Mean clustering coefficients: the role of isolated nodes and leafs on clustering measures for small-world networks.
\newblock \emph{New Journal of Physics}, 10\penalty0 (8):\penalty0 083042, 2008.

\bibitem[Saram{\"a}ki et~al.(2007)Saram{\"a}ki, Kivel{\"a}, Onnela, Kaski, and Kertesz]{saramaki2007generalizations}
Jari Saram{\"a}ki, Mikko Kivel{\"a}, Jukka-Pekka Onnela, Kimmo Kaski, and Janos Kertesz.
\newblock Generalizations of the clustering coefficient to weighted complex networks.
\newblock \emph{Physical Review E}, 75\penalty0 (2):\penalty0 027105, 2007.

\bibitem[Loukas(2019)]{loukas2019graph}
Andreas Loukas.
\newblock What graph neural networks cannot learn: depth vs width.
\newblock \emph{arXiv preprint arXiv:1907.03199}, 2019.

\bibitem[Kim et~al.(2016)Kim, Jernite, Sontag, and Rush]{kim2016character}
Yoon Kim, Yacine Jernite, David Sontag, and Alexander Rush.
\newblock Character-aware neural language models.
\newblock \emph{Proceedings of the AAAI Conference on Artificial Intelligence}, 30\penalty0 (1), Mar. 2016.
\newblock \doi{10.1609/aaai.v30i1.10362}.
\newblock URL \url{https://ojs.aaai.org/index.php/AAAI/article/view/10362}.

\bibitem[Xue et~al.(2022)Xue, Barua, Constant, Al-Rfou, Narang, Kale, Roberts, and Raffel]{xuebyt52022}
Linting Xue, Aditya Barua, Noah Constant, Rami Al-Rfou, Sharan Narang, Mihir Kale, Adam Roberts, and Colin Raffel.
\newblock Byt5: Towards a token-free future with pre-trained byte-to-byte models.
\newblock \emph{Transactions of the Association for Computational Linguistics}, 10:\penalty0 291--306, 03 2022.
\newblock ISSN 2307-387X.
\newblock \doi{10.1162/tacl_a_00461}.
\newblock URL \url{https://doi.org/10.1162/tacl\_a\_00461}.

\bibitem[Clark et~al.(2022)Clark, Garrette, Turc, and Wieting]{clark2022Canine}
Jonathan~H. Clark, Dan Garrette, Iulia Turc, and John Wieting.
\newblock Canine: Pre-training an efficient tokenization-free encoder for language representation.
\newblock \emph{Transactions of the Association for Computational Linguistics}, 10:\penalty0 73--91, 01 2022.
\newblock ISSN 2307-387X.
\newblock \doi{10.1162/tacl_a_00448}.
\newblock URL \url{https://doi.org/10.1162/tacl\_a\_00448}.

\bibitem[Zhang et~al.(2015)Zhang, Zhao, and LeCun]{zhang2015character}
Xiang Zhang, Junbo Zhao, and Yann LeCun.
\newblock Character-level convolutional networks for text classification.
\newblock In C.~Cortes, N.~Lawrence, D.~Lee, M.~Sugiyama, and R.~Garnett, editors, \emph{Advances in Neural Information Processing Systems}, volume~28. Curran Associates, Inc., 2015.
\newblock URL \url{https://proceedings.neurips.cc/paper_files/paper/2015/file/250cf8b51c773f3f8dc8b4be867a9a02-Paper.pdf}.

\bibitem[O'Shea and Nash(2015)]{o2015introduction}
Keiron O'Shea and Ryan Nash.
\newblock An introduction to convolutional neural networks.
\newblock \emph{arXiv preprint arXiv:1511.08458}, 2015.

\bibitem[Brody et~al.(2022)Brody, Alon, and Yahav]{brody2022attentive}
Shaked Brody, Uri Alon, and Eran Yahav.
\newblock How attentive are graph attention networks?, 2022.

\bibitem[Veličković et~al.(2018)Veličković, Cucurull, Casanova, Romero, Liò, and Bengio]{veličković2018graph}
Petar Veličković, Guillem Cucurull, Arantxa Casanova, Adriana Romero, Pietro Liò, and Yoshua Bengio.
\newblock Graph attention networks, 2018.

\bibitem[Gehring et~al.(2017)Gehring, Auli, Grangier, Yarats, and Dauphin]{gehring2017convolutional}
Jonas Gehring, Michael Auli, David Grangier, Denis Yarats, and Yann~N Dauphin.
\newblock Convolutional sequence to sequence learning.
\newblock In \emph{International conference on machine learning}, pages 1243--1252. PMLR, 2017.

\bibitem[Clevert et~al.(2016)Clevert, Unterthiner, and Hochreiter]{clevert2016fast}
Djork-Arné Clevert, Thomas Unterthiner, and Sepp Hochreiter.
\newblock Fast and accurate deep network learning by exponential linear units (elus), 2016.

\bibitem[Kiranyaz et~al.(2021)Kiranyaz, Avci, Abdeljaber, Ince, Gabbouj, and Inman]{KIRANYAZ2021107398}
Serkan Kiranyaz, Onur Avci, Osama Abdeljaber, Turker Ince, Moncef Gabbouj, and Daniel~J. Inman.
\newblock 1d convolutional neural networks and applications: A survey.
\newblock \emph{Mechanical Systems and Signal Processing}, 151:\penalty0 107398, 2021.
\newblock ISSN 0888-3270.
\newblock \doi{https://doi.org/10.1016/j.ymssp.2020.107398}.
\newblock URL \url{https://www.sciencedirect.com/science/article/pii/S0888327020307846}.

\bibitem[Maas et~al.(2011)Maas, Daly, Pham, Huang, Ng, and Potts]{maas2011Learning}
Andrew~L. Maas, Raymond~E. Daly, Peter~T. Pham, Dan Huang, Andrew~Y. Ng, and Christopher Potts.
\newblock Learning word vectors for sentiment analysis.
\newblock In \emph{Proceedings of the 49th Annual Meeting of the Association for Computational Linguistics: Human Language Technologies}, pages 142--150, Portland, Oregon, USA, June 2011. Association for Computational Linguistics.
\newblock URL \url{http://www.aclweb.org/anthology/P11-1015}.

\bibitem[Pang and Lee(2004)]{pang2004asentiment}
Bo~Pang and Lillian Lee.
\newblock A sentimental education: Sentiment analysis using subjectivity summarization based on minimum cuts.
\newblock \emph{arXiv preprint cs/0409058}, 2004.
\newblock URL \url{http://www.cs.cornell.edu/people/pabo/movie-review-data/}.

\bibitem[Gulli(2004)]{antonio2004agnews}
Antonio Gulli.
\newblock Ag's corpus of news articles, 2004.
\newblock URL \url{http://www.di.unipi.it/~gulli/AG_corpus_of_news_articles.html}.

\bibitem[McAuley and Leskovec(2013)]{mcauley2013hidden}
Julian McAuley and Jure Leskovec.
\newblock Hidden factors and hidden topics: understanding rating dimensions with review text.
\newblock In \emph{Proceedings of the 7th ACM conference on Recommender systems}, pages 165--172, 2013.

\bibitem[Ionescu and Butnaru(2019)]{ionescu2019vector}
Radu~Tudor Ionescu and Andrei~M Butnaru.
\newblock Vector of locally-aggregated word embeddings (vlawe): A novel document-level representation.
\newblock \emph{arXiv preprint arXiv:1902.08850}, 2019.

\bibitem[Huang et~al.(2023)Huang, Qin, Zhou, Zhu, Liu, and Shao]{huang2023normalization}
Lei Huang, Jie Qin, Yi~Zhou, Fan Zhu, Li~Liu, and Ling Shao.
\newblock Normalization techniques in training dnns: Methodology, analysis and application.
\newblock \emph{IEEE transactions on pattern analysis and machine intelligence}, 45\penalty0 (8):\penalty0 10173--10196, 2023.

\bibitem[Mikolov et~al.(2013)Mikolov, Sutskever, Chen, Corrado, and Dean]{mikolov2013distributed}
Tomas Mikolov, Ilya Sutskever, Kai Chen, Greg~S Corrado, and Jeff Dean.
\newblock Distributed representations of words and phrases and their compositionality.
\newblock \emph{Advances in neural information processing systems}, 26, 2013.

\bibitem[Sanh et~al.(2019)Sanh, Debut, Chaumond, and Wolf]{sanh2019distilbert}
Victor Sanh, Lysandre Debut, Julien Chaumond, and Thomas Wolf.
\newblock Distilbert, a distilled version of bert: smaller, faster, cheaper and lighter.
\newblock \emph{arXiv preprint arXiv:1910.01108}, 2019.

\end{thebibliography}

\end{document}